\documentclass[preprint,12pt]{elsarticle}

\usepackage{multirow}
\usepackage{amssymb}
\usepackage{hyperref}
\usepackage{booktabs}%
\usepackage{xcolor}

\usepackage{subcaption}

\journal{Heliyon}

\begin{document}

\begin{frontmatter}

\title{End-to-End Data Quality-Driven Framework for Machine Learning in Production Environment}    

\author[1]{Firas Bayram}
\affiliation[1]{organization={Department of Mathematics and Computer Science, Karlstad University}, 
            city={Karlstad},
            postcode={651 88}, 
            country={Sweden}}

\author[1,2]{Bestoun S. Ahmed}
\affiliation[2]{organization={American University of Bahrain}, 
            city={Riffa},
            country={Bahrain}}

\author[3]{Erik Hallin}
\affiliation[3]{organization={Uddeholms AB}, 
            city={Hagfors},
            postcode={683 33}, 
            state={Värmlands län},
            country={Sweden}}

\begin{abstract}

This paper introduces a novel end-to-end framework that efficiently integrates data quality assessment with machine learning (ML) model operations in real-time production environments. While existing approaches treat data quality assessment and ML systems as isolated processes, our framework addresses the critical gap between theoretical methods and practical implementation by combining dynamic drift detection, adaptive data quality metrics, and MLOps into a cohesive, lightweight system. The key innovation lies in its operational efficiency, enabling real-time, quality-driven ML decision-making with minimal computational overhead. We validate the framework in a steel manufacturing company’s Electroslag Remelting (ESR) vacuum pumping process, demonstrating a 12\% improvement in model performance (R² = 94\%) and a fourfold reduction in prediction latency. By exploring the impact of data quality acceptability thresholds, we provide actionable insights into balancing data quality standards and predictive performance in industrial applications. This framework represents a significant advancement in MLOps, offering a robust solution for time-sensitive, data-driven decision-making in dynamic industrial environments. 
\end{abstract}

\begin{highlights}
\item Real-time framework integrates data quality assessment with drift detection in MLOps
\item Improves industrial ML performance with 12\% higher accuracy and $R^2$ of 94\%
\item Reduces prediction latency by four times in time-sensitive production systems
\item Model-agnostic design enhances flexibility across diverse industrial ML use cases
\end{highlights}


\end{frontmatter}

\section{Introduction}
\label{sec:intro}

Artificial intelligence (AI) and machine learning (ML) are transforming industries by enabling companies to derive insights from data, which is essential for informed decision-making. Several fields, such as healthcare, finance, cybersecurity, and software reliability engineering, have embraced these technologies to drive innovation and efficiency \cite{kuhar2024overview}. These domains are characterized by dynamic environments where conditions continuously evolve. For instance, in healthcare, during the COVID-19 pandemic, ML systems had to adapt to shifting epidemiological trends and evolving treatment protocols while handling complex interdependencies among health parameters \cite{chitla2024multivariate}. Similarly, in software reliability engineering, ML models must continuously adjust to changing system behaviors, usage patterns, and newly emerging failure modes \cite{samal2024neural, samal2024enhancing}. This evolutionary nature of real-world applications highlights the importance of adaptive ML systems that can maintain performance despite changing conditions.

Despite these advances, much of the literature has historically focused on refining algorithms rather than on enhancing data quality \cite{whang2023data}. However, recent studies demonstrate that data quality can have a more substantial impact on ML performance than the choice of algorithm itself \cite{moran2022important}. This insight has sparked a paradigm shift from model-centric to data-centric AI \cite{hamid2022model}, emphasizing the optimization of processing methods and the training of ML models on high-quality data—especially in industrial applications where massive volumes of data are processed rapidly \cite{rosati2023knowledge}.
In data-centric approaches, one effective method for ensuring high data quality during system development is the introduction of an \textit{acceptability threshold} \cite{loshin2010practitioner}. This threshold defines a minimum standard that data must meet to be deemed suitable for training ML models, thereby ensuring that only data surpassing these predefined criteria are utilized \cite{batini2009methodologies}. To determine this threshold, data quality is assessed using quantitative methods that evaluate various dimensions—such as accuracy, completeness, consistency, and timeliness—and combine them into a comprehensive data quality score \cite{sidi2012data, ridzuan2024review}. Each dimension focuses on a specific aspect of data quality, providing a diversified evaluation that supports robust decision-making across dynamic environments.

While high-quality data is crucial for accurate decision-making, the process of monitoring and maintaining data quality imposes a significant burden on ML systems, particularly in real-time environments \cite{mcgilvray2021executing}. This burden arises from the continuous assessment and validation of incoming data against predefined quality standards, which can strain system resources and complicate management efforts \cite{jayasankar2021survey}. 
Current approaches often involve computationally intensive processes that can introduce significant delays, potentially rendering the ML predictions obsolete in time-sensitive applications. For instance, Google’s TFX \cite{polyzotis2019data} employs batch-oriented data validation, which, while thorough, is not optimized for real-time processing. Similarly, DaQL \cite{ehrlinger2019daql} faces scalability issues in dynamic environments, as its rule-based quality filters require extensive computational resources for large-scale data streams. These limitations highlight the need for lightweight frameworks that minimize latency, ensuring timely decision-making in applications where immediate actions are necessary \cite{mazumder2024dataperf}.

\begin{figure*}
    \centering
    \begin{subfigure}[b]{0.49\linewidth}
        \centering
        \includegraphics[width=\linewidth]{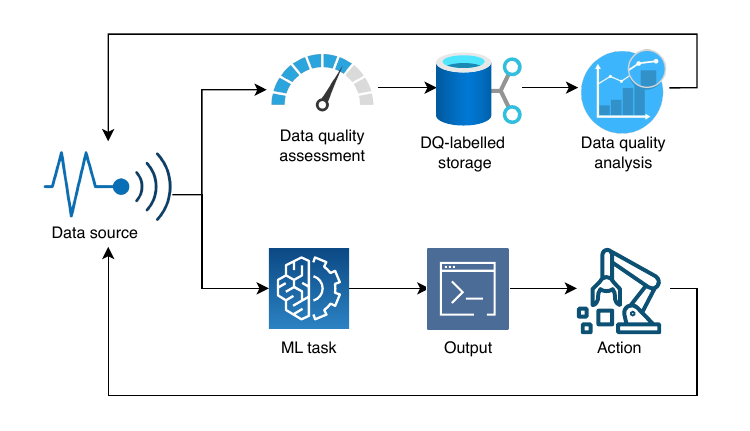}
        \caption{Traditional approach: Data quality and ML systems as separate entities.}
        \label{fig:comparison_a}
    \end{subfigure}
    \hfill
    \begin{subfigure}[b]{0.49\linewidth}
        \centering
        \includegraphics[width=\linewidth]{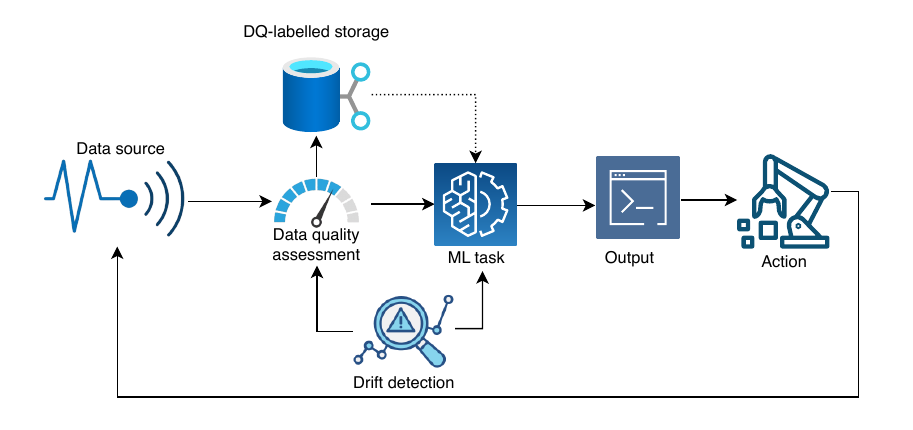}
        \caption{Proposed approach: End-to-end integration of data quality and ML systems.}
        \label{fig:comparison_b}
    \end{subfigure}
    \caption{Comparison of ML system architectures. (a) Traditional approach with offline data quality assessment. (b) Proposed integrated approach with continuous data quality monitoring and drift detection.}
    \label{fig:comparison}
\end{figure*}

Machine Learning Operations (MLOps) has emerged as a key approach to handling ML system operations in deployment environments \cite{kreuzberger2023machine}.  MLOps integrates DevOps and data engineering principles to optimize large-scale deployment, monitoring, and management of ML models \cite{testi2022mlops}. These processes facilitate iterative development, enabling the continuous update and improvement of ML models in response to changing data and conditions \cite{tamburri2020sustainable}. To optimize the efficiency of deploying and managing ML systems, addressing the challenges associated with data quality assessment in production contexts within MLOps systems is imperative. At the abstract level, as illustrated in Fig. \ref{fig:comparison_a}, data quality assessment has traditionally been treated as a separate offline task due to the complexity overhead involved. However, integrating data quality assessment directly into the system would streamline this process. Ideally, as presented in Fig. \ref{fig:comparison_b}, data quality assessment should complement ML algorithms, with drift detection acting as a dynamic bridge to ensure continuous adaptation to changing conditions. This integration is essential for the success of AI systems in real-world applications \cite{jakubik2024data}. The resulting symbiotic relationship enables a self-evolving system that automatically adjusts to data distribution shifts while maintaining quality standards, leading to more effective and reliable task handling within industrial environments.

Despite the recognized importance of data quality and the emergence of MLOps practices, there remains a significant gap in the integration of adaptive, real-time data quality assessment within operational ML systems. Many existing solutions either focus on offline data quality checks or implement simplistic, rule-based quality filters that may not capture the nuanced aspects of data quality required for complex ML applications \cite{juddoo2015overview}. Furthermore, the dynamic nature of industrial environments, where data distributions can shift rapidly, poses additional challenges to maintaining model performance and reliability over time.

This paper addresses these critical gaps by proposing a novel end-to-end framework that integrates adaptive data quality assessment with MLOps practices in real-time production systems. Using MLOps principles, such as continuous integration and deployment (CI/CD) pipelines, which automate the testing, integration, and deployment of machine learning models, the proposed system incorporates an adaptive quantitative data quality assessment into the ML model lifecycle. It also employs change detection mechanisms to identify deviations in data distributions, automatically triggering the retraining and deployment processes of the model. This approach ensures ML models' continuous robustness and adaptability to changing data conditions, enhancing their performance and reliability in dynamic production environments. The framework is designed to be efficient and lightweight, prioritizing the minimization of computation overhead and enabling timely decision-making in industries where time-sensitive decisions are essential. By enabling adaptive and efficient ML systems capable of thriving in the face of evolving data landscapes and stringent operational requirements, this framework represents a significant advancement in the field. The results of applying our framework in real-world scenarios have shown enhanced performance without a significant increase in overhead, confirming its effectiveness in practical environments.

The rest of this paper is organized as follows: Section \ref{sec:background} provides general background on the topics that form the foundation of this paper and reviews the related work. Section \ref{sec:framework} presents the proposed framework and its constituent elements. Section \ref{sec:results} details the implementation and experimental results in the industrial use case. Section \ref{sec:conclusion} summarizes the conclusions drawn from the study.

\section{Background and Related Work}
\label{sec:background}

The effectiveness of ML systems in real-world applications hinges mainly on two critical factors: the quality of data used for model training and the operational management of these models in production environments. This section explores these foundational aspects and reviews the related work that addresses the challenges and solutions in these areas.

\subsection{Key Concepts}
\label{sec:concepts}
This study deals with two essential topics: data quality assessment and MLOps. The integration between these topics guarantees the development of robust ML systems that can function effectively in operational settings.

\subsubsection{Data Quality Assessment}
Data quality assessment is fundamental to ensure that the data used in ML models are accurate, reliable, and fit for purpose. High-quality data plays a crucial role in building robust and effective ML models, as their performance and reliability depend directly on the quality of the data on which they are trained. The process of data quality assessment involves evaluating several critical attributes to determine the suitability of the data for analysis and decision-making \cite{watts2009data}. 

Two main methodologies are commonly used in data quality assessment, quantitative and qualitative assessments \cite{zaveri2016quality}:

\begin{enumerate}
    \item \textbf{Quantitative assessment:} These assessments involve using scales or metrics to measure various aspects of data quality. Quantitative evaluations provide an in-depth understanding of data characteristics. By quantifying data quality, we can objectively assess its fitness for analysis and decision-making. Quantitative assessments offer clarity and consistency in industrial contexts, allowing us to track system progress over time \cite{cai2015challenges}.
    \item  \textbf{Qualitative Assessments:} Qualitative assessments focus on data's inherent characteristics and subjective elements, often referred to as data profiling \cite{liu2009encyclopedia}. Data profiling involves examining data to understand its structure, patterns, and anomalies. While qualitative assessments lack the precision of quantitative metrics, they provide valuable insights into data behavior.
\end{enumerate}

In this study, we focus on quantitative assessment for industrial processes due to the need for objective, scalable, and standardized metrics across diverse aspects in real-time decision-making. Industries rely on established frameworks such as International Organization for Standardization (ISO) 8000 for data quality management, Six Sigma's Define-Measure-Analyze-Improve-Control (DMAIC) methodology, and Total Data Quality Management (TDQM), which provide structured evaluation methods using predefined measurement scales \cite{talburt2015entity}. This quantitative approach, based on standardized measurements and numerical scales, enhances the consistency in quality evaluation across different industrial processes. Additionally, it enables precise comparison of quality metrics over time and across different system components while facilitating automated decision-making through clear numerical thresholds and benchmarks. These characteristics make quantitative assessment particularly suitable for real-time industrial applications where objective, repeatable measurements are crucial for operational efficiency \cite{bertossi2020data}.

\begin{figure}
    \centering
    \includegraphics[width=1\linewidth]{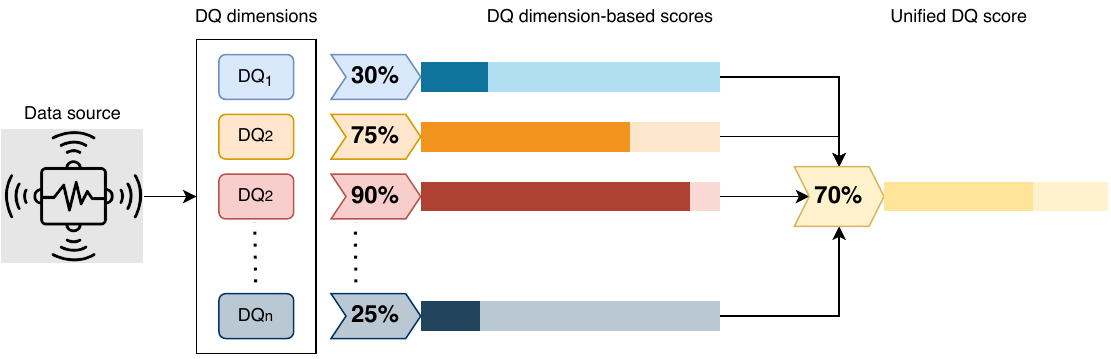}
    \caption{Data quality scoring process across multiple data quality dimensions (DQ).}
    \label{fig:dqa}
\end{figure}

To implement such quantitative assessment effectively, ensuring data quality requires a systematic evaluation across multiple dimensions. This complex, multi-faceted process involves evaluating various data quality dimensions to determine the reliability of collected data. Each dimension provides insight into different aspects of the characteristics of the data \cite{pipino2014developing}. The selection of data quality dimensions is dynamic and is often driven by the specific objectives and requirements of the processed use case. Fig. \ref{fig:dqa} illustrates the key steps in calculating a comprehensive data quality score for the collected data, starting with identifying the data quality dimensions tailored to the use case application \cite{taleb2016big}. Subsequently, individual data quality scores are computed based on each dimension. Finally, a unified data quality score is derived to represent the overall quality of the collected data.

\subsubsection{Change Point Detection}
Change point detection, or drift detection, is a statistical technique used to identify points in time-series data where the underlying statistical properties shift significantly \cite{aminikhanghahi2017survey}. These changes can be abrupt or gradual, and detecting them is vital for maintaining the robustness of ML systems' performance, especially in dynamic industrial environments. By identifying these shifts, industries can adapt their models and processes to ensure continued accuracy and reliability \cite{truong2020selective}.

In the context of industrial applications, change point detection is crucial for monitoring the processes that generate data. In manufacturing, for example, sudden changes in sensor data could indicate equipment malfunctions or deviations from standard operating conditions. Detecting these changes promptly allows for rapid intervention, reducing downtime, and maintaining production quality \cite{choi2021deep}. Furthermore, this continuous monitoring helps identify real-time data-related issues, enabling timely corrective actions. For example, if a data source starts generating anomalous readings due to sensor failure or environmental changes, change point detection can trigger alerts for further investigation and remediation \cite{ahmadzadeh2018change}.

\begin{figure}
    \centering
    \includegraphics[width=0.8\linewidth]{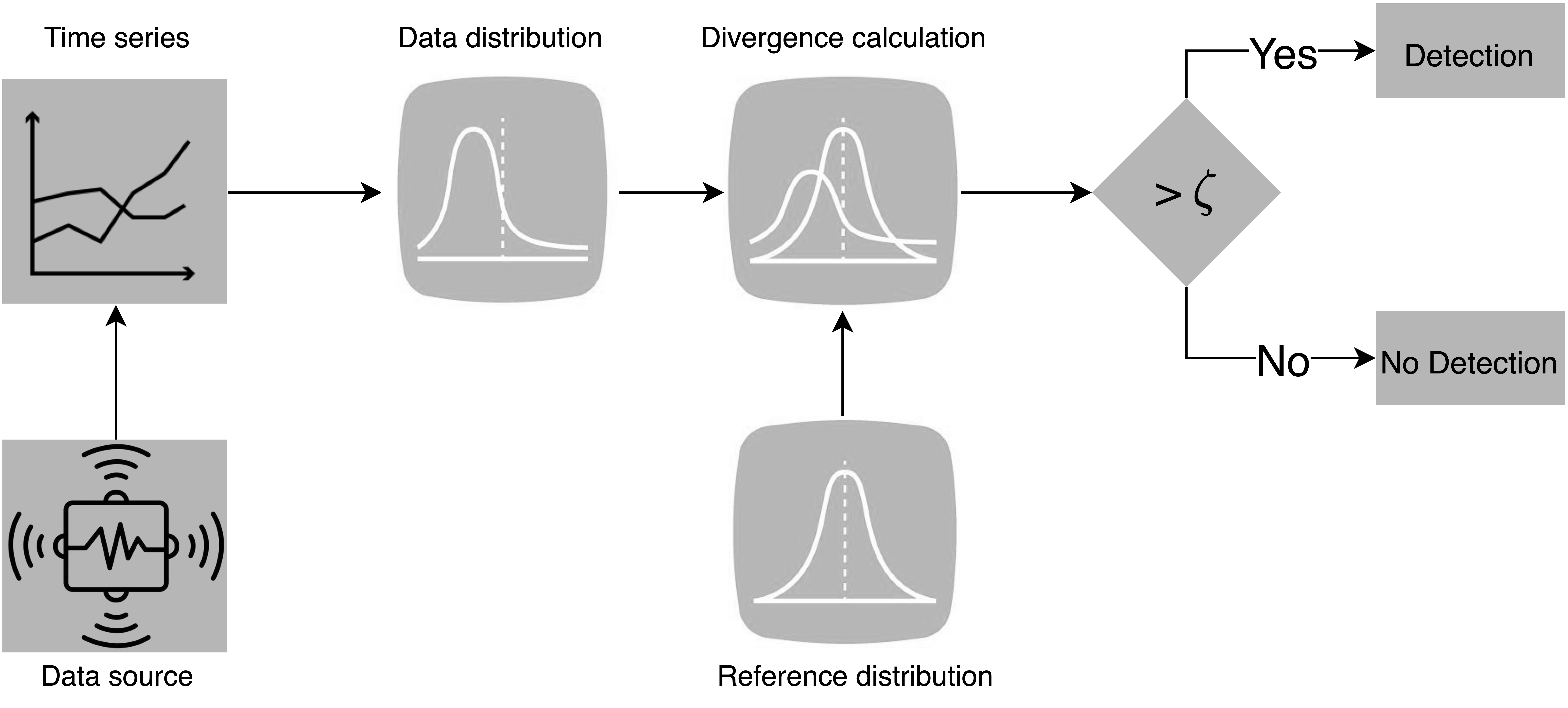}
    \caption{Change point detection mechanism.}
    \label{fig:cpd}
\end{figure}

Change point detection methods often involve assessing the divergence between two distributions: one representing the reference and the other representing the recent distribution, to verify if a change has occurred \cite{tartakovsky2014sequential}. Implementing change point detection involves segmenting the time series data, computing divergence measures for these segments, and applying statistical tests to identify significant shifts. Hypothesis testing is commonly used to determine whether there is a significant difference between segments of the time-series data. As shown in Fig. \ref{fig:cpd}, divergence measures $D_t$ like Kullback-Leibler (KL) Divergence and Jensen-Shannon (JS) Divergence quantify the disparity between probability distributions $P$ and $Q$ at the timestamp $t$ \cite{liu2013change}:
\begin{equation}
D_{t} = \delta(P \parallel Q).
\end{equation}
If $D_{t} > \zeta$, where $\zeta$ is a pre-specified threshold, a change is detected, signifying a significant shift in the underlying data distribution. Consequently, the system should be updated to adapt to the prevailing conditions \cite{lu2018learning}.

\subsubsection{Machine Learning Operations (MLOps)}
Machine Learning Operations (MLOps) is a set of practices that aim to streamline and orchestrate the deployment, monitoring, and maintenance of ML models in production \cite{john2021towards}. MLOps integrates the development of the ML system (Dev) with the operations of the ML system (Ops), ensuring that ML models can be deployed reliably and maintained efficiently. The core principles of MLOps include continuous integration and continuous deployment (CI/CD) of ML models, continuous monitoring and logging, artifact versioning and reproducibility \cite{symeonidis2022mlops}.

CI/CD pipelines automate the process of integrating changes and deploying models to production environments, ensuring the rapid deployment of new features, updates, and fixes. Continuous monitoring and logging track the performance and health of ML models in production by capturing metrics related to model predictions, data quality, and system performance, which helps detect anomalies and diagnose issues \cite{ruf2021demystifying}. Artifacts versioning is necessary for managing different versions of system artifacts used for training and inference, ensuring scalability and traceability of ML experiments. Reproducibility involves maintaining detailed records of data, code, model parameters, and configurations, allowing others to replicate results and validate findings.

In industrial settings, the adoption of MLOps becomes indispensable for effectively managing real-time applications and systematically integrating ML models into operational workflows. Specifically, MLOps allows for system artifacts to be continuously improved through quick updates and retraining to adapt to changing data and conditions. This adaptability is particularly critical in dynamic industrial environments, where quick adjustments are essential to maintain model accuracy and reliability \cite{subramanya2022devops}. Moreover, by following MLOps practices, ML models can be scaled and managed effectively, providing a reliable system for handling the lifecycle of ML models in production environments.

\subsection{Related Work}
\label{sec:related}

The integration of data quality assessment with machine learning operations presents a significant challenge in production environments, particularly when real-time performance is crucial. Traditionally, data quality assessment and the development of high-quality ML models in operational settings have been studied separately in the literature~\cite{liu2022survey}, often due to the high complexity and added overhead in integrating these components. Recent advances in MLOps practices have sparked interest in integrating data quality assessment methodologies into ML model development pipelines~\cite{rangineni2023analysis}. Our analysis of existing literature reveals three primary research streams: theoretical frameworks for data quality assessment, practical MLOps implementations, and attempts to bridge these domains.

In the theoretical domain, recent work has established foundational frameworks for data quality assessment in ML contexts. According to a recent survey by Singh~\cite{singh2023systematic}, existing studies primarily focus on discussions and recommendations about data quality frameworks in the context of MLOps, lacking empirical experimentation and practical implementation. Seedat et al.~\cite{seedat2022dc} introduced the DC-Check framework, which provides a comprehensive checklist for data-centric considerations throughout the ML pipeline. While this framework offers valuable guidelines, it lacks mechanisms for real-time implementation and doesn't address the computational overhead challenges critical for production environments. Similarly, Jain et al.~\cite{jain2020overview} presented insights from IBM Research\footnote{https://research.ibm.com/} on data quality analysis for ML applications, but their approach primarily focuses on offline analysis, leaving real-time assessment challenges unaddressed.

On the practical implementation front, several systems have attempted to operationalize data quality assessment in ML pipelines. Google's TFX data validation system~\cite{polyzotis2019data} represents a significant advancement in production-scale data validation, incorporating components for batch and inter-batch validation. However, its architecture prioritizes thoroughness over real-time processing speed, making it less suitable for time-sensitive industrial applications. The DaQL library~\cite{ehrlinger2019daql} offers a more targeted approach for industrial ML applications, but its evaluation revealed significant performance limitations that restrict its applicability in real-time scenarios.  Recent work by Nain \textit{et al.}\cite{nain2024packmasnet} demonstrates how deep learning solutions for quality inspection can be effectively deployed on resource-constrained edge devices, addressing the challenges of model retraining and storage optimization in Industry 5.0 settings. Similarly, Nain \textit{et al.} \cite{nain2023novel} propose mechanisms for handling dynamic manufacturing environments through continual learning-based model retraining at the edge. These works highlight the growing importance of efficient MLOps practices in resource-limited industrial settings. However, while these approaches effectively address model deployment and retraining challenges, they lack comprehensive integration of data quality assessment within their edge computing frameworks, particularly for optimizing ML inference model performance.

A separate survey conducted a temporal mapping analysis of data quality requirements within ML development pipelines~\cite{priestley2023survey}, highlighting the significance of addressing these requirements in various stages of the ML data pipeline. The work by Chen \textit{et al.}~\cite{chen2021data} comes closest to addressing real-time data quality challenges, proposing a transfer-learning approach to improve data quality and identifying key quality attributes such as comprehensiveness, correctness, and variety. However, their solution focuses primarily on offline learning scenarios and doesn't address the specific challenges of continuous quality monitoring in production environments. This limitation is particularly significant given the increasing importance of real-time decision-making in industrial applications.

Our review of the literature reveals three critical gaps in current research:

\begin{enumerate}
    \item \textbf{Real-time Performance:} Existing frameworks primarily focus on offline or batch processing, failing to address the computational efficiency required for real-time operations in production environments.
    
    \item \textbf{Integration Overhead:} Current solutions either introduce significant computational overhead or require substantial modifications to existing ML pipelines, making them impractical for production deployment.
    
    \item \textbf{Industrial Applicability:} Most frameworks lack validation in industrial settings where real-time performance and reliability are crucial, such as manufacturing processes.
\end{enumerate}

Our proposed framework addresses these limitations by introducing a lightweight, real-time data quality assessment mechanism that integrates seamlessly with existing ML operations. Unlike previous approaches that treat data quality assessment as a separate process, our framework incorporates continuous quality monitoring without compromising system performance. This innovation enables real-time decision-making while maintaining high prediction accuracy, as demonstrated in our industrial case study of the steel manufacturing ESR process.

\section{Proposed Data Quality-Driven ML Framework}
\label{sec:framework}

\begin{figure}
    \centering
    \includegraphics[width=1\linewidth]{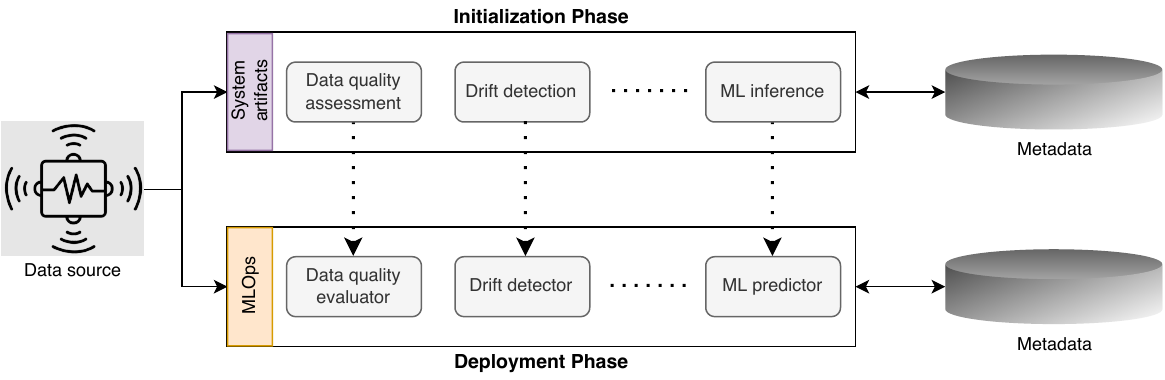}
    \caption{Phases of the proposed framework.}
    \label{fig:phases}
\end{figure}

This section outlines the proposed framework designed to ensure that decisions made by ML systems are driven by high data quality standards. The framework integrates a data quality assessment method that supports the training and maintenance of ML models, all within the context of MLOps practices. This approach ensures efficiency of production and systematic maintainability of the system. As shown in Fig. \ref{fig:phases}, the framework consists of two major phases: the \textit{Initialization Phase}, where system artifacts are built, and the \textit{Deployment Phase}, where these artifacts are utilized to continuously operate and adapt in the production environment. Each phase is supported by \textit{metadata} stores, which act as central repositories for configurations used throughout the system's lifecycle.  Specifically, metadata stores can hold information about the drift detection method, the configuration of the data quality scoring module, and the training settings of the data quality ML model.

\subsection{System Artifacts Initialization Phase}
\begin{figure}
    \centering
    \includegraphics[width=1\linewidth]{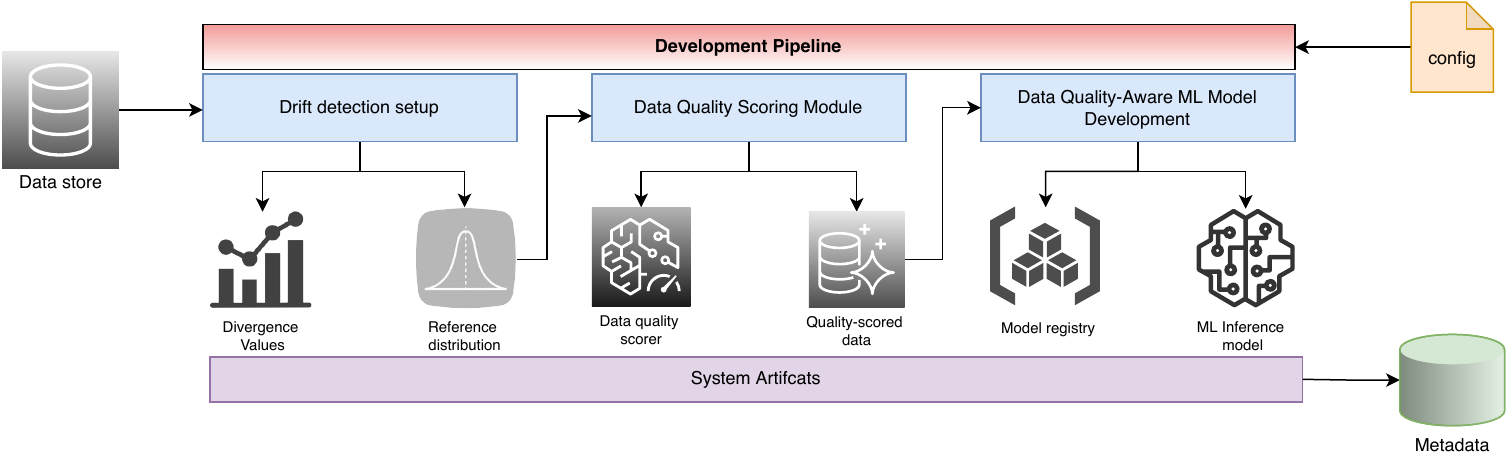}
    \caption{Initialization phase pipeline and associated system artifacts.}
    \label{fig:development}
\end{figure}

In this phase, the warm-start process is initiated. Specifically, the main task is to initialize and prepare the artifacts of the system necessary for the deployment of the ML system. This encompasses a series of steps to establish and develop the foundational elements required for the subsequent deployment and operations of the ML models within the framework. As illustrated in Fig. \ref{fig:development}, three key elements form the basis of this development pipeline of the system artifacts: the drift detection mechanism setup, the data quality scoring module and the development of the data quality-aware ML model. The drift detection mechanism initiates the creation of meta-information crucial for monitoring changes during production and calculating the relevant data quality scores. Concurrently, the data quality scoring module assesses and scores the incoming data across various quality dimensions. Data quality-aware ML model development focuses on creating and training ML models that inherently account for data quality scores. These artifacts will be pushed and integrated into the production environment to operationalize the entire ML system. 

\subsubsection{Drift Detection Setup}\label{sec:drift_setup}
The drift detection setup is a critical component of the system initialization phase, developed to monitor and detect changes in data distribution over time. Traditionally, drift detection occurs when the magnitude of change exceeds predefined thresholds. However, within dynamic industrial environments, establishing an optimal threshold for detecting drift presents a formidable challenge \cite{agrahari2022concept}. Industrial data streams inherently possess complex and fluctuating statistical properties, making the conventional static threshold approach unviable. Therefore, recent studies suggest adopting adaptive thresholds that dynamically adjust to accommodate the evolving system dynamics and underlying processes \cite{liu2022concept}.

\begin{figure}
    \centering
    \includegraphics[width=0.75\linewidth]{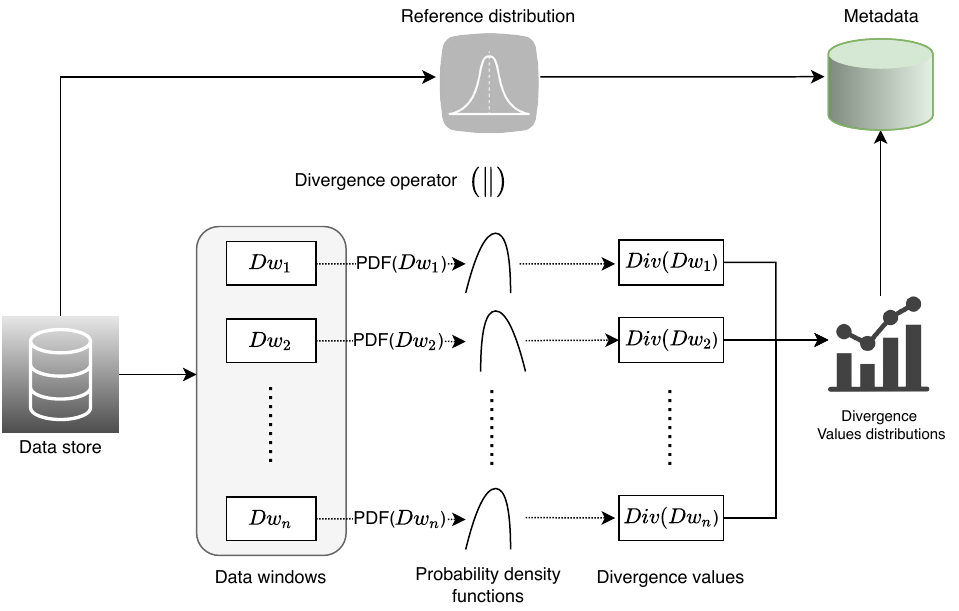}
    \caption{Schematic representation of the drift detection setup process.}
    \label{fig:dd}
\end{figure}

To address the dynamic characteristic of industrial data in real time, we utilize our previously developed method for drift detection \cite{bayram2023lstm}. Unlike traditional approaches that depend on fixed drift magnitude thresholds, our method dynamically adjusts to changing conditions without needing predefined thresholds. The schematic workflow of the adaptive approach within the drift detection setup is illustrated in Fig. \ref{fig:dd}. The process involves segmenting the dataset into data windows. The probability density functions (PDFs) for these windows are then constructed, and the divergence between these PDFs and a reference distribution is derived from the baseline data. Afterward, we construct a distribution based on the calculated divergence values, which evolves as more data windows are collected. In the production phase, this distribution enables us to assess the extent to which the change in the observed data window deviates from the normal change values.

\subsubsection{Data Quality Scoring Module}
\label{sec:dqsops}
The data quality scoring module assesses the quality of incoming data across various data quality dimensions. As shown in Fig. \ref{fig:dqa}, the module evaluates the collected data window based on the pre-defined set of data quality dimensions and assigns a unified score that reflects the overall data quality. This score indicates the quality measure for the data window, which is crucial for determining the suitability of the data for training the ML models. We selected data quality dimensions that are especially relevant for our industrial application. These dimensions are described below:
\begin{enumerate}
     \item \textbf{Accuracy:} This dimension evaluates the extent to which recorded data aligns with its intended real-world representation. The accuracy score is determined by the percentage of anomalous data found in the window, represented as \(Accuracy = \frac{NAV}{N}\), where \(NAV\) denotes the total number of anomalous values and \(N\) is the window size.
    \item \textbf{Completeness:} Evaluating the comprehensiveness of the data window, completeness analysis detects and quantifies missing values within the data window. The completeness score is determined as \(Completeness = \frac{NNV}{N}\), where \(NNV\) represents the count of missing values (e.g., "NA").
    
    \item \textbf{Consistency:} This dimension evaluates whether observed values adhere to defined integrity constraints, ensuring that data values fall within suitable ranges. The consistency score is calculated based on the number of consistent values, given as \(Consistency = \frac{NCV}{N}\), where \(NCV\) is the count of consistent values.
    
    \item \textbf{Timeliness:} Describing the relevance of data for specific tasks, timeliness assesses the currency of observed data achieved through a goodness-of-fit test \cite{grosso2023fast}, such as a two-sample Kolmogorov-Smirnov test, which compares current data with a reference distribution. The timeliness score is computed using the Kolmogorov-Smirnov test statistic \(KS = \max_{1 \leq i \leq N}|\hat{F_1}(Z_i) - \hat{F_2}(Z_i)|\), where \(Z\) is the combined sample of two independent random samples \(X\) and \(Y\).
    
    \item \textbf{Skewness:} This dimension explores the distribution deviation of the data window from a reference distribution, employing the Jensen-Shannon Divergence (JSD) value to quantify the dissimilarity between their distributions. JSD is calculated as \(\mathrm{JSD}(P\| Q) = H\left(\frac{P+Q}{2}\right) - \frac{H(P)+H(Q)}{2}\), where \(H\) denotes Shannon's entropy. JSD is preferred for its constrained behavior within the interval \([0,1]\) \cite{lionis2021rssi}.
\end{enumerate}
Following the evaluation of these dimensions, the principal component analysis (PCA) technique is applied to combine the individual scores \cite{teh2020sensor}. Subsequently, these combined scores are standardized to generate a comprehensive unified score, providing a holistic measure of overall data quality.

Given the substantial computational resources necessary for calculating these scores, particularly in real-world applications where data arrives in real-time and large volumes, a cost-effective ML-based approach is commonly used to manage this task \cite{schelter2018automating}. This approach involves utilizing ML models trained on annotated ground truth data to label incoming data based on its quality attributes. Our previous research introduced the Data Quality Scoring Operations (DQSOps) framework \cite{bayram2023dqsops}, a novel ML-based approach designed to score data quality. The DQSOps framework is integrated into MLOps systems to streamline the data quality scoring process. To increase robustness, the DQSOps framework also incorporates data mutation to induce synthetic data quality-related issues, enabling the model to better generalize in the presence of real-world data anomalies. The framework demonstrated superior computational efficiency compared to traditional methods while maintaining high levels of predictive accuracy. Furthermore, to further improve the capabilities of data quality scoring in dynamic environments, we developed an adaptive data quality scoring method \cite{bayram2024adaptive}. This approach dynamically adjusts the data quality scores based on evolving data patterns and prevalent conditions, addressing the challenges posed by real-time, non-stationary environments.

As shown in Fig. \ref{fig:development}, the main outcomes of this module include two key artifacts: the ML model responsible for scoring the incoming data windows in production and the annotated data containing quality scores, which are used to train the ML inference model. The ML model is implemented using XGBoost, a choice validated through our previous work in data quality assessment. XGBoost has demonstrated efficient data handling capabilities and superior performance in industrial data streams, making it particularly suitable for real-time applications \cite{bayram2023dqsops, bayram2024adaptive}. The model is developed by training on annotated ground truth data, which is prepared by evaluating data quality across various dimensions, as described earlier. Once the model achieves a specified level of accuracy in quality score prediction, it is deployed for production use.

\subsubsection{Data Quality-Aware ML Model Development}
\begin{figure}
    \centering
    \includegraphics[width=0.8\linewidth]{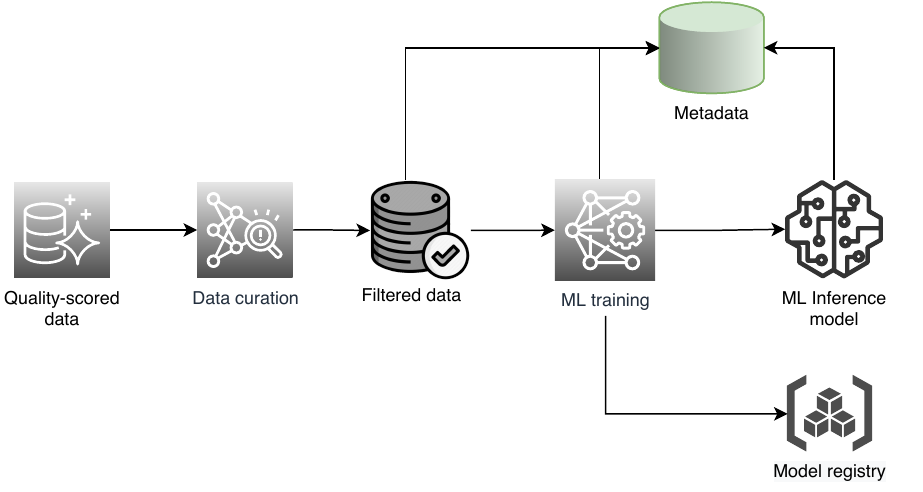}
    \caption{Overview of the data-quality aware ML model development process.}
    \label{fig:qml}
\end{figure}

The data quality-aware ML model development process focuses on creating ML models that integrate data quality assessments into their training processes. As depicted in Fig. \ref{fig:qml}, this phase directly incorporates data quality scores to enhance the robustness and performance of the model. The process begins with the preparation of training data, where the raw data is annotated with quality scores generated by the data quality scoring module. The data is then curated to retrieve a training dataset that maintains a required quality standard. This step involves removing data whose quality score falls below an acceptable threshold, ensuring that only high-quality data are used for model training.

After preparing and curating the data, the next phase involves training the ML models. These models use filtered data that have passed through the data curation process. This approach promotes more accurate and reliable predictions by mitigating the impact of low-quality data. Once the models are trained, they undergo a validation process to assess their predictive performance before being deployed into production. To ensure reproducibility and traceability, a model registry systematically tracks and logs all activity related to the model, storing different versions of the ML model parameters and training settings.

\subsection{Deployment Phase}
\begin{figure}
    \centering
    \includegraphics[width=1\linewidth]{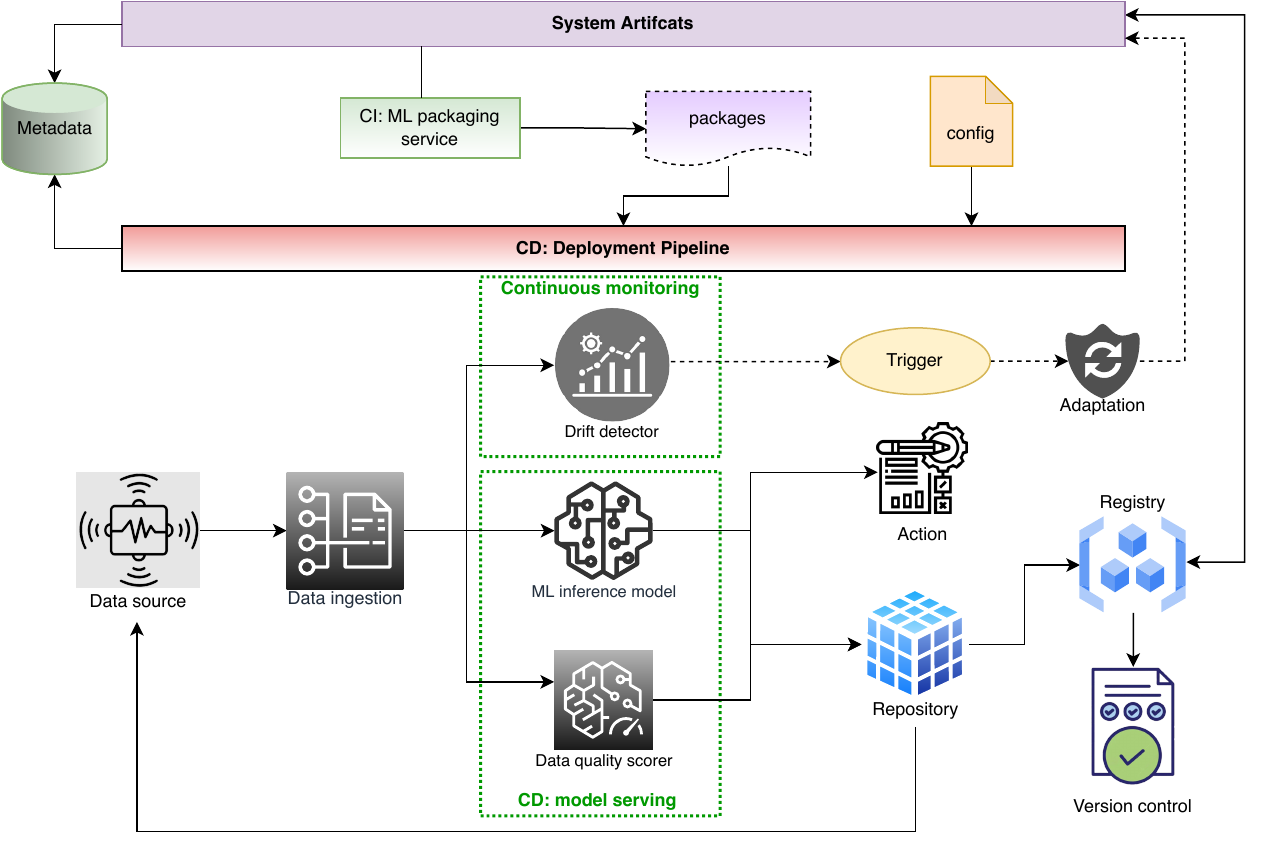}
    \caption{CI/CD pipeline of the deployment phase.}
    \label{fig:deployment}
\end{figure}

In the deployment phase, the system artifacts initialized during the initialization phase are used to continuously operate and evolve in the production environment. The main components of this phase include continuous integration (CI) for packaging, continuous deployment (CD) for serving, and continuous monitoring of the system. As shown in Fig. \ref{fig:deployment}, the deployment pipeline starts by ingesting the data window from the data source. Once the data window is collected, it is fed into two main services: continuous monitoring and the ML model-serving service. Continuous monitoring checks for any significant changes in the data to activate adaptation signals. Meanwhile, the ML model-serving service ensures that trained models are readily available for real-time inference requests. This iterative process of monitoring and adaptation is the cornerstone of our adaptive ML system, enabling it to dynamically respond to evolving conditions in the production environment.

\subsubsection{Continuous Drift Monitoring}\label{sec:con_drift}
Continuous drift monitoring service involves real-time detection of shifts in data distribution that could affect model performance. This service continuously ingests live production data and compares its statistical properties with the reference artifacts produced in the initialization phase. As shown in Fig. \ref{fig:deployment}, alerts are triggered to prompt adaptation signals to update the relevant artifacts when the drift detector identifies a significant change. This adaptation process involves retraining ML models for inference and scoring. Additionally, artifacts such as data quality scores and reference data distributions are updated to reflect current conditions and ensure the system remains up-to-date with the prevalent conditions. Each update introduces new versions, enabling traceability and facilitating analysis, maintenance, and debugging efforts.

\subsubsection{ML Models Serving}
The ML models in this service are designed to facilitate the transition from development to deployment, ensuring that trained models are readily accessible for real-time prediction requests. Once deployed, these models process the production data and make predictions or decisions using the latest versions of the models. The framework encompasses two main types of ML models: a data quality scoring model and an ML inference model. The data quality scoring model, which operates as a regressor, assesses and scores production data. The ML inference model makes predictions relevant to the industrial task, supporting and optimizing industrial processes for appropriate actions. All predictions are stored in a central repository alongside corresponding versions of the ML model artifacts. This centralized storage allows for further analysis and performance validation of the models. All predictions are generated in real time, facilitating optimal performance and decision-making within the industrial context.

\section{Implementation Details and Results}
\label{sec:results}

The proposed data quality-driven ML framework has been integrated into a comprehensive AI system designed to improve decision-making processes in industrial environments. To evaluate its effectiveness, the framework was implemented in a real-world scenario: the Electroslag Remelting (ESR) vacuum pumping process in Uddeholm\footnote{https://www.uddeholm.com/en/}, a steel manufacturing company in Sweden. This AI-driven application aims to ensure high-quality steel production by monitoring and managing pressure levels. The steel manufacturing use case was selected based on its inherent characteristics. The ESR process operates under strict quality control and real-time operational constraints, making it an ideal testbed for evaluating adaptive ML frameworks. Additionally, the vacuum pumping process generates high-velocity sensor data prone to quality variations, a common challenge in modern industrial systems. Furthermore, the process requires continuous monitoring and adaptive control to ensure high-quality steel production through precise pressure level management.

The framework is designed to be model-agnostic, enabling it to support various data types and ML models across different domains. It is initially intended for deployment on machines that share similar characteristics, where a warm-up phase establishes machine-specific reference distributions and quality profiles. However, the model is also capable of generalizing across machines through parameter reinitialization, reducing the need for complete retraining. With appropriate domain-specific adjustments, the framework can be adapted to other industrial sectors that exhibit similar real-time monitoring and control requirements. These sectors commonly face challenges such as high-frequency data generation, stringent quality constraints, and the need for rapid decision-making. To validate the framework's effectiveness in addressing these common industrial challenges, we evaluated its performance based on drift detection capabilities, the impact of data quality acceptability thresholds on prediction accuracy, and real-time processing latency.

In our experiments, the baseline setup corresponds to the standard approach, which does not employ any adaptation mechanism, neither active nor passive drift detection, and relies solely on static model inference. Additionally, this baseline, used to assess the efficiency of data quality incorporation, which does not include any real-time data quality evaluation or adaptive retraining, making it a representative benchmark of traditional, non-adaptive industrial ML deployments. Improvements in predictive performance were computed relative to this baseline, using consistent evaluation settings and equivalent train-test splits across all experiments. Latency was measured from data ingestion to prediction delivery, capturing the full system pipeline under both baseline and adaptive configurations.

\subsection{Industrial Application and Experiments}
Implementing the framework within the ESR vacuum pumping process involved several key steps. The system collects and analyzes pressure data from sensors within the furnace vacuum chamber. Each vacuum pumping cycle, lasting up to 20 minutes, records pressure readings every 100 milliseconds. These readings are transmitted in real time using the Apache Kafka platform\footnote{https://kafka.apache.org/}, ensuring continuous monitoring and analysis. Each input sample is a matrix of size 1200~$\times$~$f$, where $f$ is the number of historical pump cycles collected, and 1200 corresponds to two minutes of 100ms-interval readings. The model output is a single scalar value predicting the minimum pressure expected by the end of the cycle. The training data includes both high- and low-quality data samples as part of the data quality scoring methodology, as explained in Section~\ref{sec:dqsops}, to ensure the model remains robust when exposed to low-quality input during inference.

Ground truth labels for training are derived from collecting the minimum pressure values per pump event, making this a supervised regression task. The framework monitors vacuum pump operations as part of the decision-making process. If irregularities are detected, such as poor pump operation, the system triggers alarms to alert the maintenance team. This proactive approach enables timely interventions, such as stopping the pump, to prevent inefficiencies and additional costs. The primary objective is to gradually lower the pressure values throughout the pump cycles until the desired minimum pressure is reached within the specified time period.

The ML system predicts the minimum pressure based on the first two minutes of recording each pump event to achieve this optimization. This allows us to stop improper events early, saving time and resources. The XGBoost model, with the standard Mean Squared Error (MSE) as the loss function, was used as the ML algorithm due to its efficiency and speed in industrial applications~\cite{kiangala2021effective}. While the framework maintains model-agnostic flexibility, XGBoost was particularly compelling due to its proven efficacy in industrial contexts.
Compared to alternative algorithms like Random Forests or neural networks, XGBoost demonstrates superior computational efficiency—a critical attribute for real-time inference in latency-sensitive environments. Its ability to generate interpretable feature importance scores provides engineers with valuable insights into model decision-making, which is paramount in safety-critical applications \cite{liu2022xgboost}.

Although deep learning models such as Long Short-Term Memory (LSTM) networks could potentially capture more nuanced temporal dependencies, they often require substantial computational resources and large training datasets. In contrast, XGBoost offers a more pragmatic solution for scenarios constrained by limited training samples and computational limitations, making it an ideal choice for precise and efficient predictive modeling in industrial settings \cite{giannakas2021xgboost}.

\subsection{Data Quality Thresholds and Parameter Selection}
\begin{figure}
    \centering
    \includegraphics[width=0.75\linewidth]{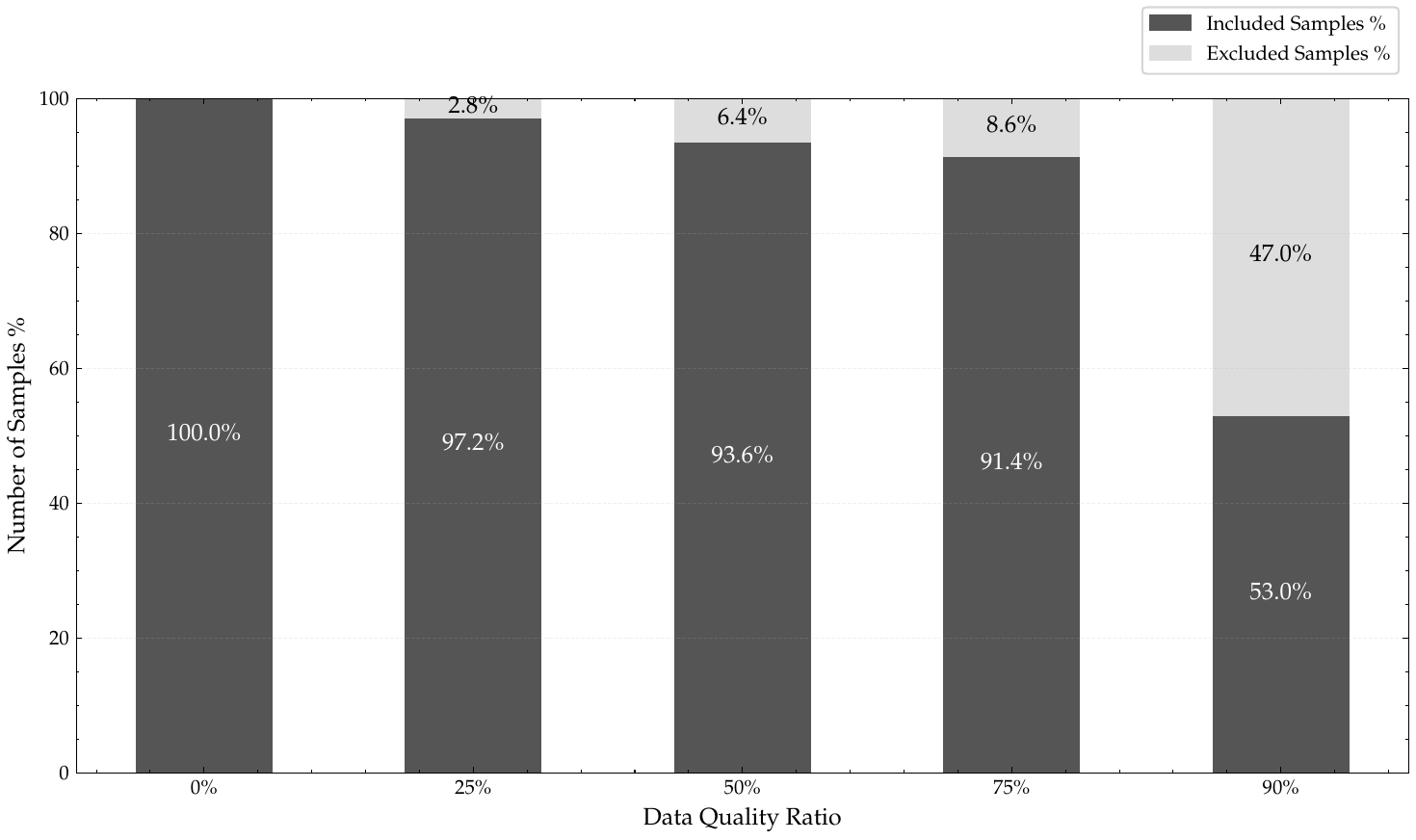}
    \caption{Data quality ratios and data sample sizes.}
    \label{fig:dq_ratio}
\end{figure}

To assess the impact of data quality on model performance, we evaluated the model using different data quality acceptability thresholds. These thresholds represent precomputed data quality scores normalized between 0 and 100 that are used to filter data to train the ML inference model. We specifically test ratios of 0, 25\%, 50\%, 75\%, and 90\% to explore how data quality affects the accuracy of the model as measured by this score. Figure \ref{fig:dq_ratio} illustrates the distribution of data quality ratios and sample sizes. A threshold of 0\% implies that the data is not filtered based on quality scores, which results in the use of all available data for training. In contrast, higher thresholds reduce the data included in the training, with 90\% filtering approximately 47\% of the full dataset.

To verify the efficiency of the adaptation method, we compare the system performance by integrating both passive and active drift detectors \cite{han2022survey}. For passive drift detection, which updates system artifacts based on predefined window sizes, we selected window sizes (w = 50, 100, 200) through extensive empirical tuning on production-derived validation data. Smaller windows enable quicker adaptation at the risk of noise sensitivity, while larger windows provide greater stability.
For active drift detection, which continuously evaluates incoming data and triggers immediate adaptations when distributional changes are detected (as explained in Section \ref{sec:con_drift}), we carefully calibrated threshold values ($\tau$ = 0.04, 0.06, 0.08). This range allows us to examine various adaptation sensitivities—lower $\tau$ values (0.04) produce more conservative adaptation behavior, while higher values (0.08) offer more responsive reactions to changes, as discussed in Section \ref{sec:drift_setup}.
Additionally, we included the standard data quality scoring method, which is not ML-based but scores the data manually according to the specified data quality dimensions.

\subsection{Predictive Performance}
Through thorough experimentation in different scenarios, we evaluated the predictive capabilities of our data quality-driven ML framework. Specifically, predictive performance was measured using Mean Absolute Error (MAE) and $R^2$ values across different data quality acceptability thresholds. The evaluation process involved applying the framework to varying levels of data quality to understand its impact on model accuracy and reliability. We set various data quality acceptability thresholds to find the best balance between data inclusion and improving model performance.

\begin{figure}[htbp]
    \centering
    
    \begin{subfigure}[b]{0.49\textwidth}
        \centering
        \includegraphics[width=\linewidth]{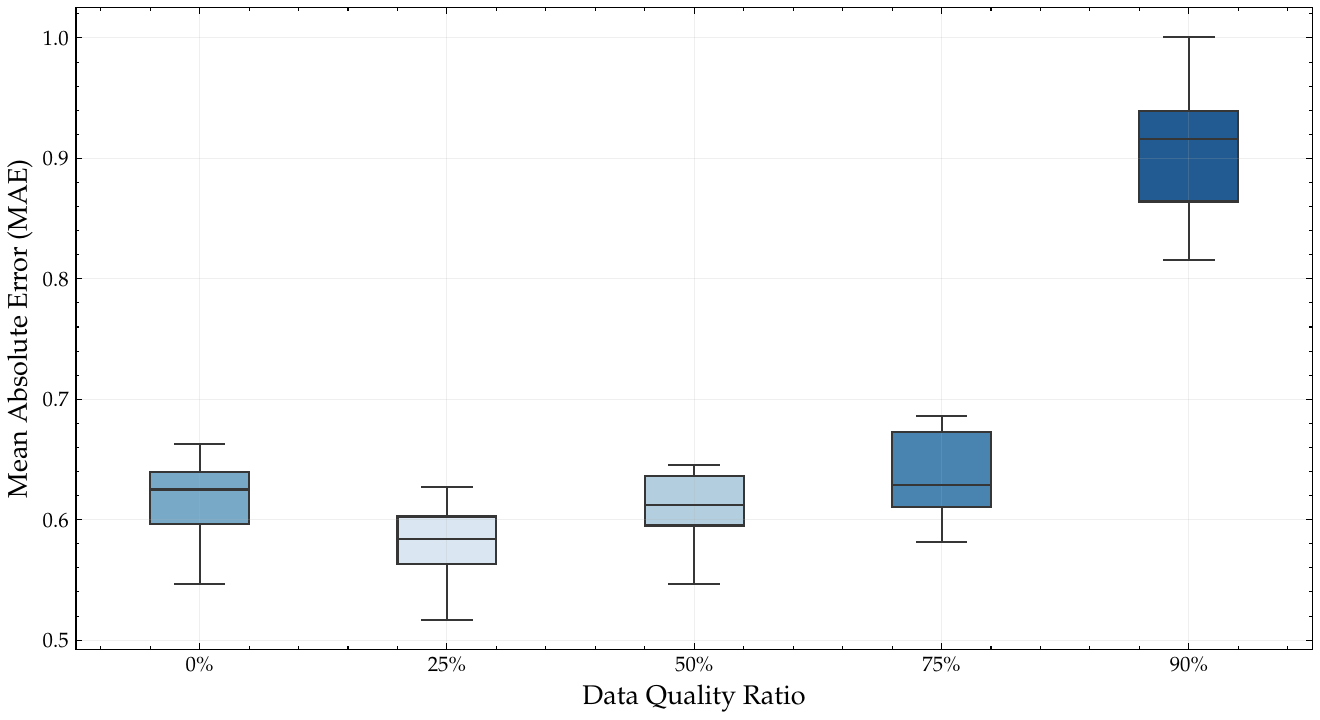}
        \caption{Distribution of MAE across various data quality ratios.}
        \label{fig:mae_box}
    \end{subfigure}
    \hfill
    \begin{subfigure}[b]{0.49\textwidth}
        \centering
        \includegraphics[width=\linewidth]{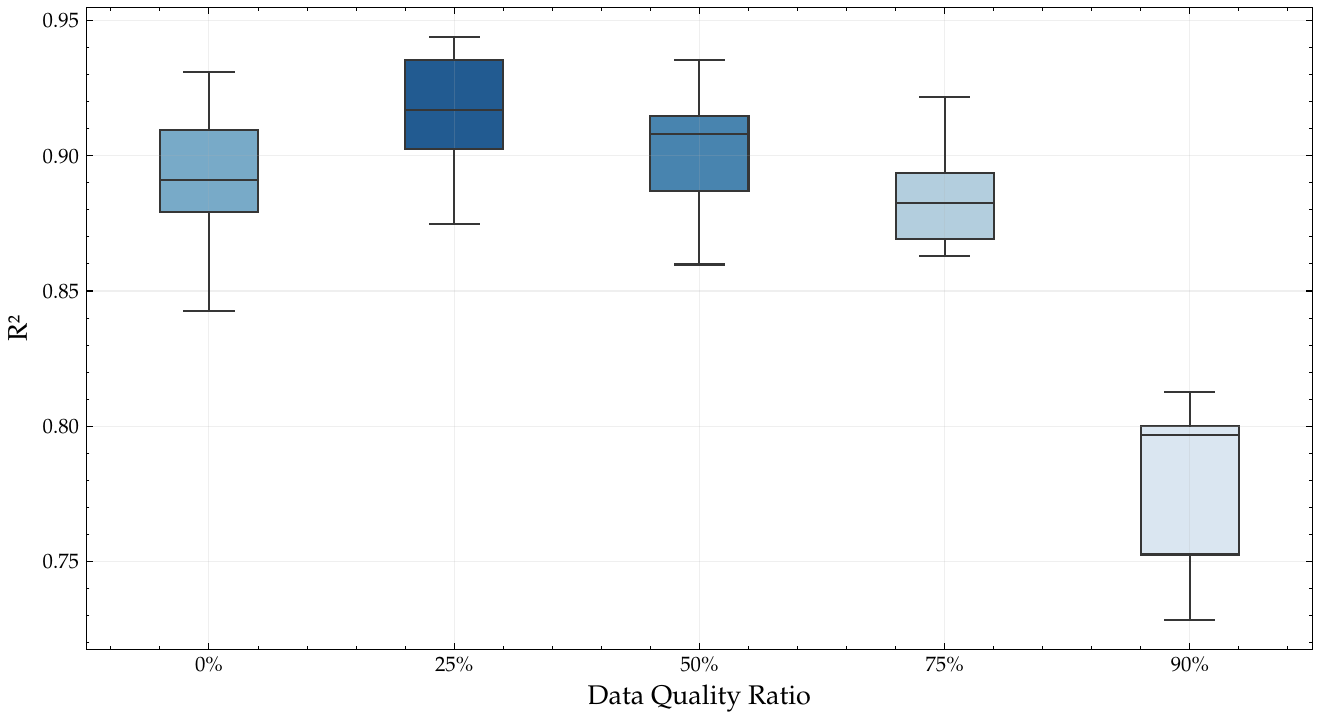}
        \caption{Distribution of $R^2$ across various data quality ratios.}
        \label{fig:r2_box}
    \end{subfigure}
    
    \caption{Performance metrics across different data quality ratios. 
             (a) Distribution of MAE. 
             (b) Distribution of $R^2$.}
    \label{fig:boxplots}
\end{figure}

As illustrated in Figures \ref{fig:mae_box} and \ref{fig:r2_box}, the ML model performance varied with the data quality used for training. Specifically, we observed a decrease in error rates as we improved the data quality of the training set to some extent. However, as more data points were filtered out, the error rates increased, indicating a loss of important information necessary for training. Our experiments revealed that moderate levels of data quality filtering, particularly the 25\% acceptability threshold, produced optimal predictive performance. This threshold effectively filtered out the most problematic data points without excessively reducing the dataset size. However, beyond this threshold, as the data quality acceptability threshold increased, the performance began to decline significantly, especially when filtering reached 90\%, revealing a trade-off between data quality and model accuracy.

\begin{figure}[htbp]
    \centering
    
    \begin{subfigure}[b]{0.49\textwidth}
        \centering
        \includegraphics[width=\linewidth]{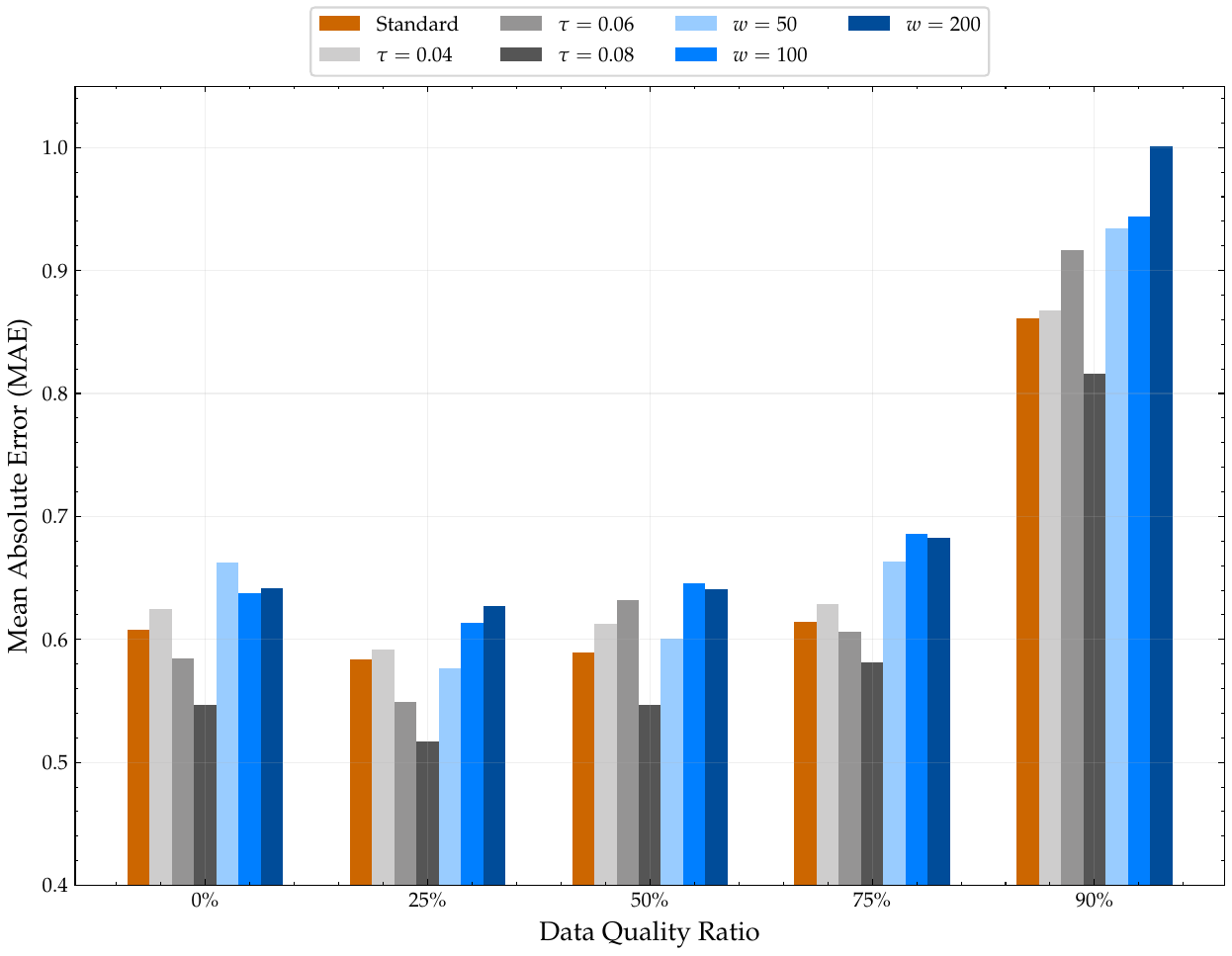}
        \caption{MAE values across different data quality acceptability thresholds.}
        \label{fig:mae}
    \end{subfigure}
    \hfill
    \begin{subfigure}[b]{0.49\textwidth}
        \centering
        \includegraphics[width=\linewidth]{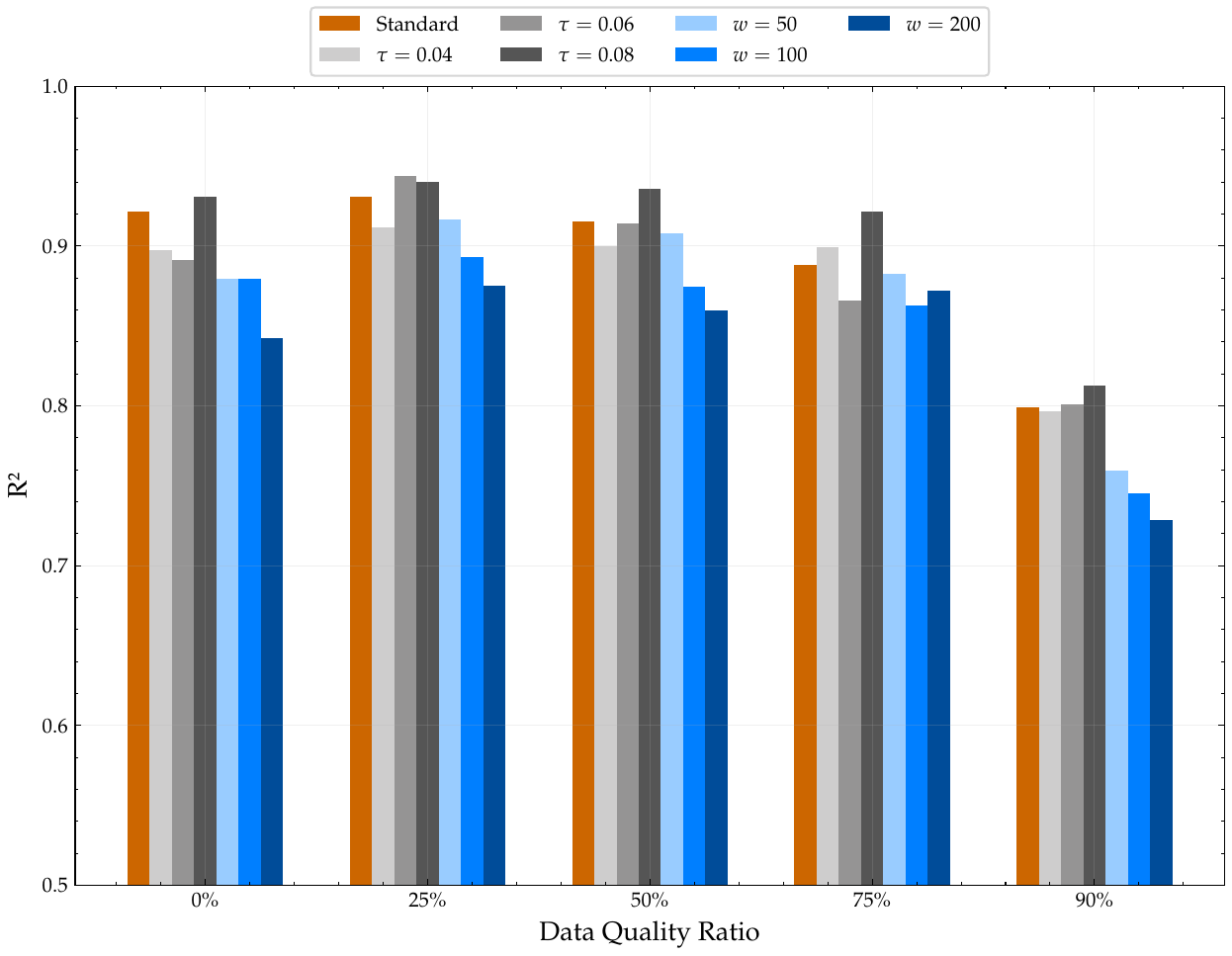}
        \caption{$R^2$ values obtained under various data quality acceptability thresholds.}
        \label{fig:r2}
    \end{subfigure}
    
    \caption{Model performance under varying data quality acceptability thresholds. 
             (a) MAE. 
             (b) $R^2$.}
    \label{fig:metrics}
\end{figure}

Moreover, we found that more adaptations resulted in better predictive performance for both active and passive approaches, as summarized in Figures \ref{fig:mae} and \ref{fig:r2}. Specifically, for the 25\% acceptability threshold, $\tau = 0.08$ for the active approach with MAE values below 0.51 and $R^2$ values around 94, respectively. Meanwhile, $w = 50$ for the passive approach achieved superior performance with an MAE below 0.58 and $R^2$ around 92. Furthermore, while generally effective, the standard approach showed slightly higher MAE values above 0.58 and $R^2$ values around 93, indicating that the active approach outperformed it in terms of predictive accuracy.

\begin{figure}
    \centering

    \begin{subfigure}[b]{\textwidth}
        \centering
        \includegraphics[width=0.85\linewidth]{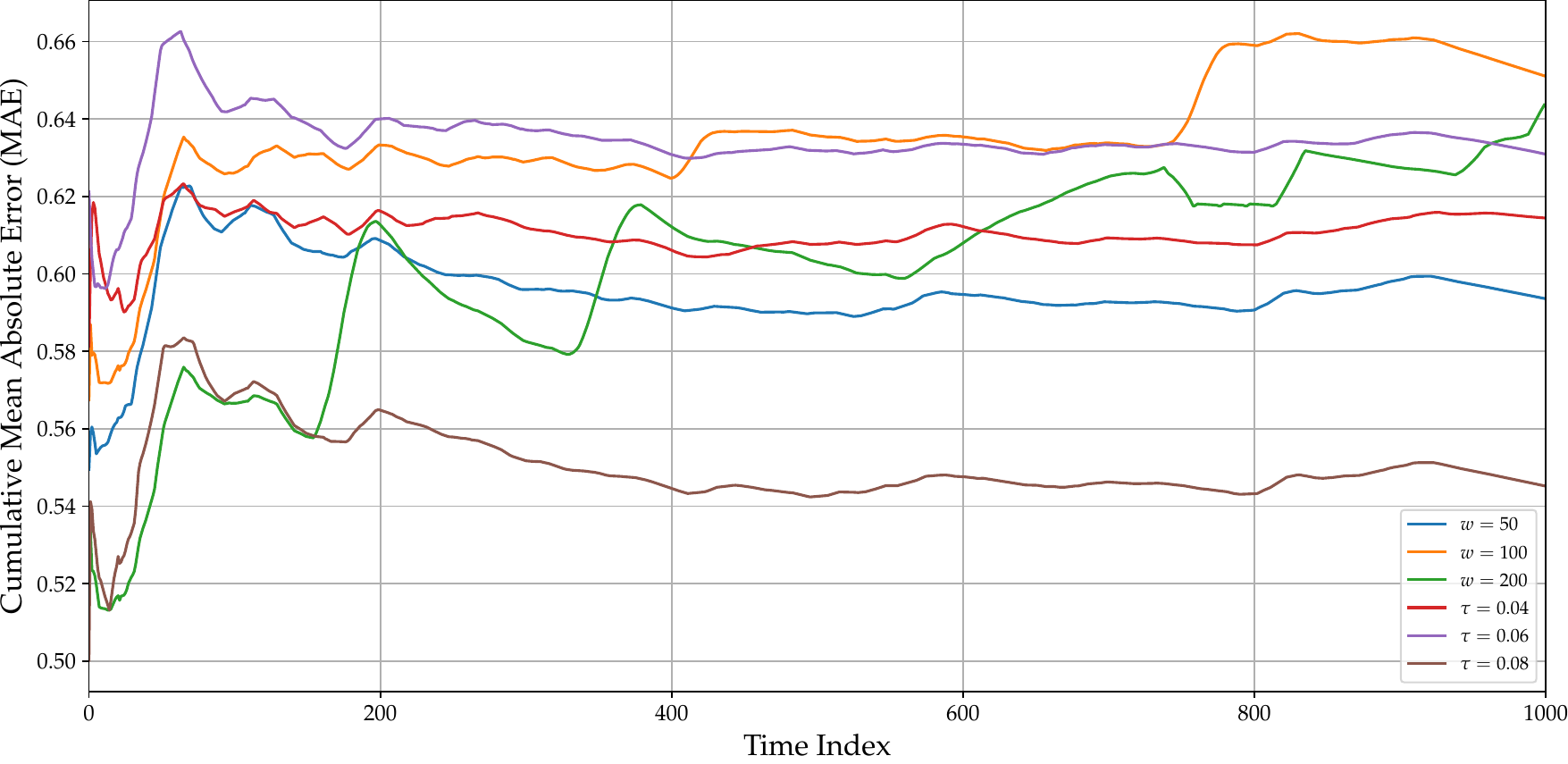}
        \caption{Temporal performance evolution under the 50\% data acceptability threshold using different adaptation strategies. (a) MAE. (b) $R^2$.}
        \label{fig:cum_mae}
    \end{subfigure}

    \vspace{0.5cm}

    \begin{subfigure}[b]{\textwidth}
        \centering
        \includegraphics[width=0.85\linewidth]{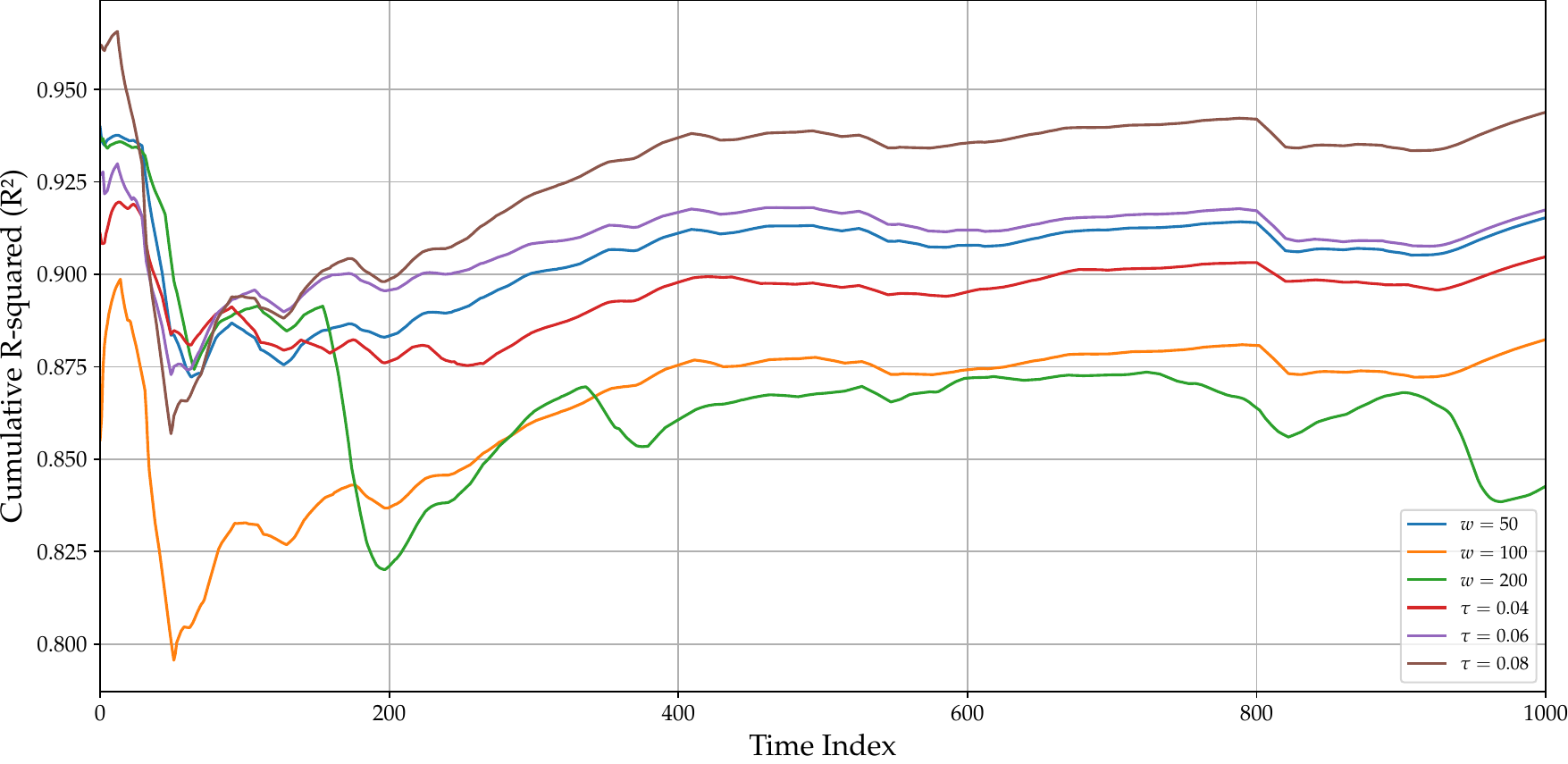}
        \caption{Temporal evolution of $R^2$ across different adaptation strategies under the 50\% acceptability threshold.}
        \label{fig:cum_r2}
    \end{subfigure}

    \caption{Performance comparison of MAE and $R^2$ over time for various adaptation strategies at the 50\% data acceptability threshold.}
    \label{fig:cumulative_metrics}
\end{figure}

Figure~\ref{fig:cumulative_metrics} shows the temporal evolution of model performance under the 50\% data quality acceptability threshold for different adaptation strategies, with Figure~\ref{fig:cum_mae} presenting MAE and Figure~\ref{fig:cum_r2} presenting $R^2$. The active adaptation strategy ($\tau=0.08$) demonstrates consistently smoother convergence, with performance metrics stabilizing earlier compared to passive approaches. Both window sizes ($w=50,200$) of the passive method exhibit more pronounced fluctuations, particularly during periods of concept drift, revealing their inherent latency in responding to distributional changes. The 50\% threshold appears to strike an effective balance - maintaining sufficient training data volume while filtering critical quality issues, as evidenced by the active method's sustained $R^2 > 0.92$ after the initial adaptation phase. This longitudinal view complements our point-in-time metrics by revealing how adaptation strategies behave differently under sustained operational conditions.

\subsection{Prediction Latency}

The prediction latency of our proposed data quality-driven ML framework is a critical factor, particularly in industrial applications requiring real-time decision-making. In this context, latency refers to the time elapsed from data ingestion to the availability of actionable predictions. The design of our framework prioritizes minimizing latency to ensure effective operation in high-stakes environments, such as our industrial use case.
To assess latency, we measured the end-to-end processing time required for the entire data pipeline, encompassing data ingestion, quality scoring, model inference, and prediction delivery.

Table \ref{tab:latency_results} compares the cumulative prediction latency, measured in seconds, for different methods of our data quality-driven ML framework across varying acceptability thresholds and adaptation settings. The standard method, which does not employ drift detection, shows high latency values starting at 801.83 seconds, decreasing to 521.97 seconds at the highest threshold. In comparison, the active methods demonstrate significantly lower latencies with varying computational impacts across configurations. With $\tau = 0.04$, the system maintains relatively low overhead, showing latencies between 147.31s to 88.53s across thresholds, while achieving effective drift detection. Increasing the adaptation sensitivity to $\tau = 0.08$ leads to higher computational costs, with latencies ranging from 193.33s to 105.50s, representing approximately a 31\% increase in base processing time. This increase correlates with the number of adaptations performed - 9 adaptations for $\tau = 0.04$ versus 20 for $\tau = 0.08$. However, despite the higher computational cost, $\tau = 0.08$ achieves superior predictive performance with MAE values below 0.51 and $R^2$ values around 94\%, suggesting that the additional computational investment yields tangible improvements in model accuracy.
The passive method exhibits intermediate performance, with a window size of $w = 50$ showing latency ranges from 286.23 to 137.75 seconds. Across all methods--standard, active, and passive--latency consistently decreases with increasing acceptability thresholds, attributable to the reduced amount of data processed as the threshold increases, allowing the framework to deliver predictions more quickly.

\begin{figure}
    \centering
    \includegraphics[width=1\linewidth]{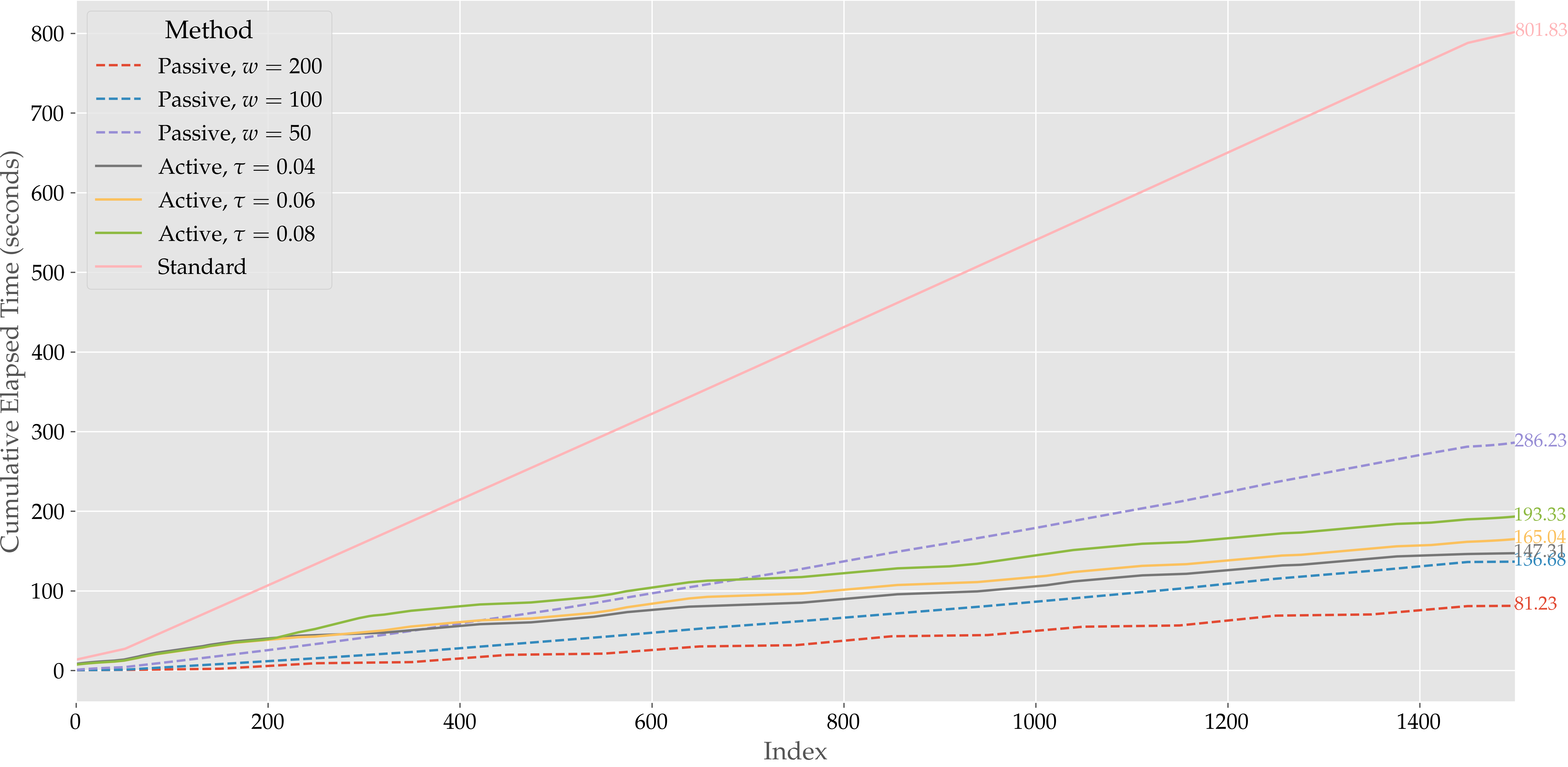}
    \caption{Prediction latency values across various data quality acceptability thresholds.}
    \label{fig:latency}
\end{figure}

In addition to the cumulative latency analysis, monitoring the prediction latency trend as the system processes data over time is crucial. Figure \ref{fig:latency} illustrates the progression of cumulative elapsed time for each method as more data flows through the system, offering insights into the efficiency of each approach. Taking the scenario in which the entire dataset is utilized with an acceptability threshold of 0, we observe a significant variation in the rate at which latency accumulates across the different methods. The standard method exhibits a rapid and nearly linear increase in latency, resulting in a maximum latency of 801.83 seconds. This sharp escalation highlights the inefficiency of the standard method, especially as data volume increases, making it unsuitable for contexts that require prompt predictions. In contrast, the active methods demonstrate a more conservative and controlled latency increase. Specifically, the active method maintains a steady, gradual increase in cumulative elapsed time. This trend shows the efficiency in managing larger datasets without experiencing a substantial latency spike, ensuring that prediction delivery remains timely. The passive methods are less conservative in their latency progression and tend to scale more noticeably with data growth. Although still more efficient than the standard method, passive methods exhibit a latency trend that more closely follows the increase in data volume, making them less efficient than active methods in real-time scenarios where minimizing latency is critical.

\begin{table}[h]
\centering
\caption{Comparison of prediction latency (in seconds) for different data quality assessment methods with varying acceptability thresholds.}
\label{tab:latency_results}
\begin{tabular}{l|c|ccccc}
\toprule
\multirow{2}{*}{\textbf{Method}} & \textbf{Number of} & \multicolumn{5}{c}{\textbf{Acceptability Thresholds}} \\ \cmidrule(lr){3-7}

\textbf{} & \textbf{adaptations} & \textbf{0} & \textbf{25} & \textbf{50} & \textbf{75} & \textbf{90} \\
\midrule
\textbf{Standard}             & -  & 801.83 & 789.55 & 780.73 & 769.25 & 521.97 \\
\textbf{Active, $\tau = 0.04$} & 9  & 147.31 & 139.84 & 132.27 & 125.12 & 88.53 \\
\textbf{Active, $\tau = 0.06$} & 14 & 165.04 & 154.73 & 144.51 & 134.28 & 94.79 \\
\textbf{Active, $\tau = 0.08$} & 20 & 193.33 & 178.07 & 163.75 & 159.50 & 105.50 \\
\textbf{Passive, $w = 50$}    & 30 & 286.23 & 262.00 & 239.25 & 217.79 & 137.75 \\
\textbf{Passive, $w = 100$}   & 14 & 136.68 & 116.75 & 117.05 & 107.35 & 78.52 \\
\textbf{Passive, $w = 200$}   & 7  & 81.23  & 76.68  & 73.80  & 70.92  & 52.03  \\
\bottomrule
\end{tabular}
\end{table}

\subsection{Data Quality and Predictive Models Performance Correlation}

\begin{figure}[ht]
    \centering
    
    \begin{subfigure}[b]{0.48\textwidth}
        \centering
        \includegraphics[width=\textwidth]{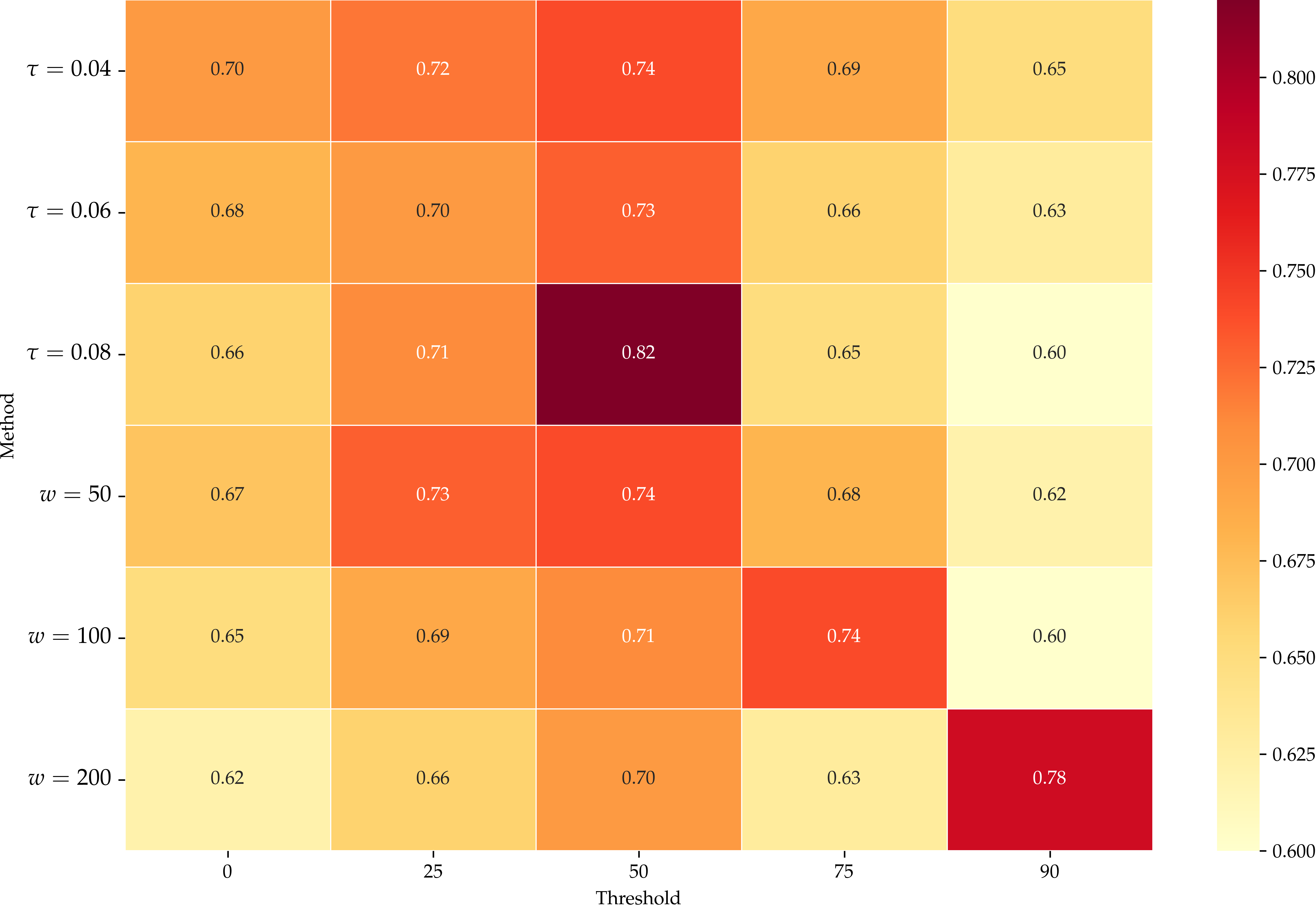}
        \caption{MAE Correlation}
        \label{fig:mae_correlation}
    \end{subfigure}
    \hfill
    \begin{subfigure}[b]{0.48\textwidth}
        \centering
        \includegraphics[width=\textwidth]{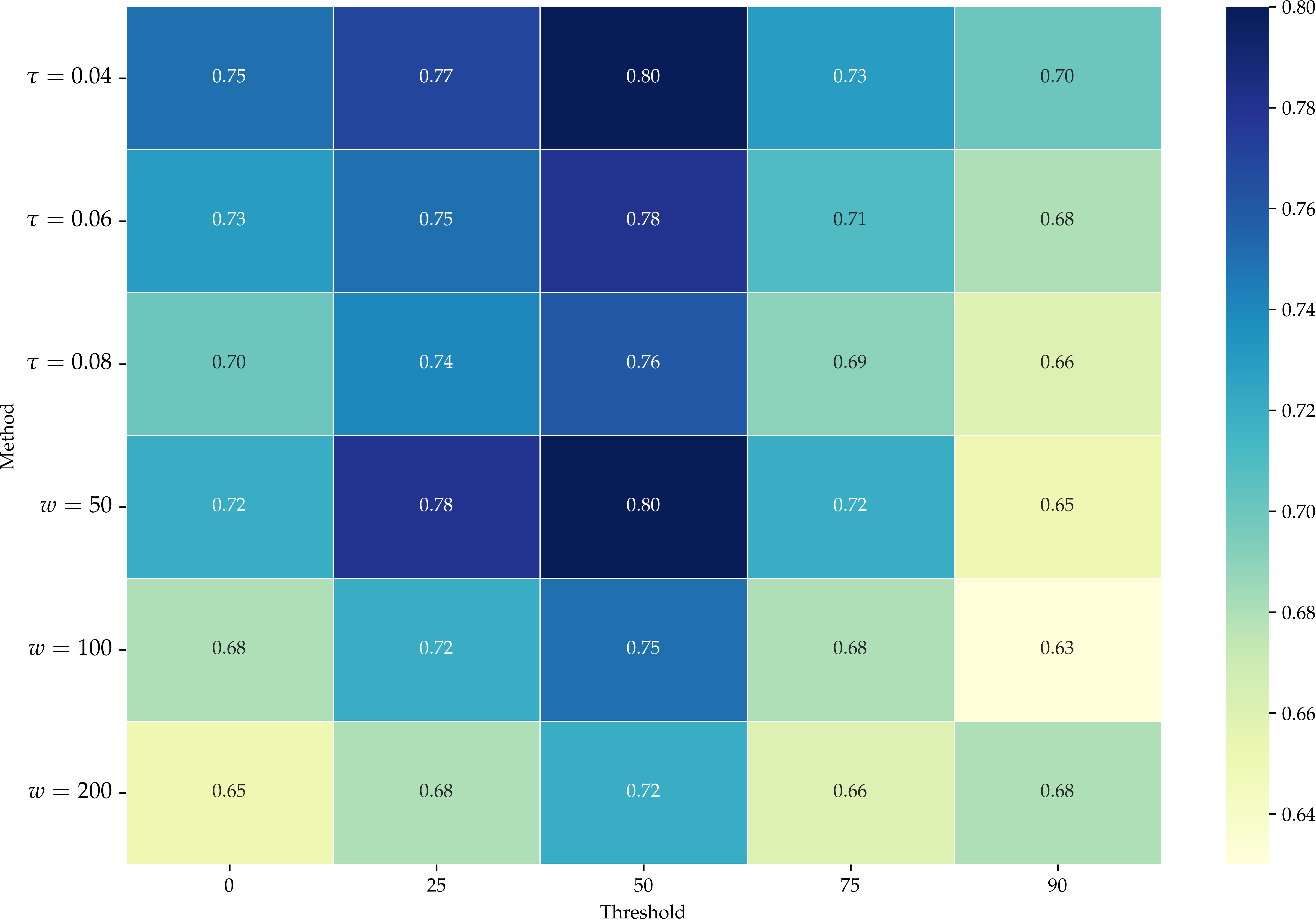}
        \caption{R2 Correlation}
        \label{fig:r2_correlation}
    \end{subfigure}

    \caption{Correlation between data quality scores and model performance across different acceptability thresholds. 
             (a) MAE. 
             (b) $R^2$.}

    \label{fig:correlation_heatmaps}
\end{figure}

To explore the relationship between the performance of data quality and inference ML models, we performed a correlation analysis of the MAE and R2 performance metrics. Figure \ref{fig:correlation_heatmaps} displays heatmaps that illustrate these correlations across various acceptability thresholds and adaptation methods.  The heatmaps reveal a generally positive correlation between data quality scores and both MAE, in Figure \ref{fig:mae_correlation}, and R2 metrics, in Figure \ref{fig:r2_correlation}, across all methods and thresholds. This strong positive relationship indicates that higher data quality generally corresponds to better model performance, thus validating the effectiveness of our data quality-driven approach. Furthermore, the strength of this correlation varies across different acceptability thresholds, with most methods showing a peak correlation at the 50\% threshold. This suggests that this mid-range threshold might represent our framework's optimal balance point for data quality assessment.

Active drift detection methods, particularly with $\tau$ = 0.08, show the strongest correlations (0.82 for MAE, 0.76 for R2) at the 50\% threshold. Passive methods with smaller window sizes ($w$ = 50, 100) perform comparably, while the larger window size ($w$ = 200) shows unique behavior with the highest correlation at the 90\% threshold for MAE.
R2 correlations are generally higher than the MAE correlations, indicating that data quality may have a stronger influence on overall model fit than on absolute prediction errors. Active and passive methods show decreased correlations at extreme thresholds (0\% and 90\%), suggesting that moderate quality criteria yield optimal results.
These findings highlight the effectiveness of our adaptive methods in aligning data quality with model performance. The analysis provides crucial insights for tuning our data quality-driven ML framework, helping to optimize the balance between quality standards and predictive performance in industrial applications.

\subsection{Fundamental Findings and Lessons Learned}

The integration of our data quality-driven ML framework within the ESR vacuum pumping process has provided several key insights and lessons that can guide future implementations in industrial environments:

\begin{enumerate}
    \item \textbf{Optimal Data Quality Threshold:} A moderate data quality acceptability threshold  produced the best predictive performance. This finding suggests that while filtering out low-quality data is beneficial, being too stringent can lead to a loss of valuable information.
    \item \textbf{Trade-offs Between Latency and Accuracy:} The results indicate a clear trade-off between reducing latency and maintaining accuracy. While higher data quality thresholds reduce latency due to smaller datasets, they also risk dropping model accuracy. Active methods with adaptive thresholds demonstrated the best balance, offering reduced latency without significant sacrifices in predictive performance.
    \item \textbf{Correlation Between Data Quality and Model Performance:} A strong correlation generally exists between the quality of data and the performance of ML models. Higher data quality tends to correlate with better predictive metrics, such as lower error rates and higher model fit. This relationship proves the importance of integrating data quality assessments throughout the ML pipeline.
    \item \textbf{Model-Agnostic Framework Design:} Designing frameworks that are model-agnostic enhances flexibility and scalability, allowing for the integration of different ML models based on specific application needs. This adaptability is particularly valuable in industrial settings where requirements may vary depending on the use case or operational constraints.
    \item \textbf{Scalability Considerations:} The latency analysis revealed that active methods scale more efficiently with increasing data volume compared to passive and standard methods. This makes active methods particularly suitable for large-scale, real-time industrial applications.
    
\end{enumerate}

\subsection{Limitations}

While the proposed framework demonstrates strong performance in the evaluated industrial use case, certain limitations highlight opportunities for future exploration:

\begin{enumerate}

\item \textbf{Extreme Data Variability:} The adaptive drift detection mechanism may require additional adaptation triggers, leading to increased computational overhead. The added complexity in detecting subtle distribution shifts could increase processing demands and potentially affect real-time performance.

\item \textbf{Sparse Datasets:} The data quality (DQ) ratio methodology may exhibit different behaviors when applied to datasets with sparse data. Since data quality is calculated through ratios, the DQ scores may require refinement for more accurate assessment in low-volume scenarios.

\item \textbf{Generalizability Across Data Types:} The current implementation is focused on structured, time-series data. Extending the framework to handle unstructured or heterogeneous data sources would enhance its applicability, particularly in complex industrial environments that involve diverse data formats.

\item \textbf{Storage Efficiency and Version Management:} Although the framework follows best practices for minimizing storage overhead—such as compressing logs, limiting metadata, and managing multiple versions of ML models and associated data quality scores remains resource-intensive. In long-running or high-frequency production environments, the cumulative storage demand can grow significantly. Future work will focus on further optimizing storage and versioning strategies to improve long-term scalability without compromising traceability and auditability.

\end{enumerate}

\section{Conclusion}
\label{sec:conclusion}
This study presents a novel data quality-driven ML framework that efficiently integrates data quality assessment with ML model operations in real-time production environments. The framework implementation in the ESR vacuum pumping process demonstrated its ability to enhance decision-making processes in industrial settings while maintaining efficiency. The adaptive approach of the framework, particularly the active drift detection method, showed superior performance in terms of predictive accuracy and latency compared to standard methods. Furthermore, the system demonstrated significant reductions in prediction latency compared to standard methods, especially when using active drift detection approaches. This improvement in latency is crucial for real-time industrial applications where quick decision-making is essential. Additionally, a strong positive correlation between data quality scoring and ML inference model performance metrics (MAE and R2) was observed, validating the effectiveness of the data quality-driven approach, validating the framework's effectiveness and highlighting the importance of high data quality standards in ML systems.
The framework's ability to dynamically adapt to changing data distributions ensures consistent model performance in non-stationary industrial environments. This adaptability is particularly valuable in industrial settings where conditions can change rapidly and unexpectedly. These outcomes emphasize the significance of integrating data quality assessment within MLOps practices to improve the robustness and reliability of ML systems in industrial applications.
Future work could focus on several important areas to further expand the framework. Exploring the capability of the framework to accommodate a wider range of data types and industrial processes would demonstrate its flexibility and applicability to various contexts. Furthermore, integrating new real-time monitoring tools and feedback mechanisms could also offer more granular insights into system performance, enabling continuous improvement and responsiveness. Addressing these areas will ensure that the framework remains robust and effective as industrial environments and data complexities evolve.

\section*{Acknowledgment}
This work has been funded by the Knowledge Foundation of Sweden (KKS) through the Synergy Project AIDA - A Holistic AI-driven Networking and Processing Framework for Industrial IoT (Rek:20200067).

\section*{Data and Code Availability}  
The experimental data used in this study were obtained from an industrial partner and are subject to confidentiality agreements. As a result, the data cannot be shared publicly. However, the code can be shared upon reasonable request.

\bibliographystyle{elsarticle-num-names}
\bibliography{bibliography}

\end{document}